\pdfoutput=1
\documentclass[10pt]{article}

\usepackage[letterpaper,textwidth=5.5in,textheight=9in,top=1in,
            headheight=12pt,headsep=25pt,footskip=30pt]{geometry}
\usepackage[utf8]{inputenc}
\usepackage[T1]{fontenc}
\usepackage{times}
\usepackage[round,authoryear]{natbib}
\setcitestyle{citesep={;},aysep={,},yysep={;}}
\usepackage{amsmath}
\usepackage{amsfonts}
\usepackage{amssymb}
\usepackage{bbm}
\usepackage{nicefrac}
\usepackage{microtype}
\usepackage{xcolor}
\usepackage{array}
\usepackage{booktabs}
\usepackage[most]{tcolorbox}
\usepackage{enumitem}
\usepackage{caption}      
\usepackage{subcaption}
\usepackage{graphicx}
\usepackage{float}
\usepackage{placeins}     
\usepackage{makecell}
\usepackage{listings}
\usepackage{url}
\usepackage[colorlinks=true,linkcolor=blue!50!black,citecolor=blue!50!black,urlcolor=blue!50!black]{hyperref}

\makeatletter
\renewcommand{\section}{\@startsection{section}{1}{\z@}%
  {-2.0ex plus -0.5ex minus -0.2ex}{1.5ex plus 0.3ex minus 0.2ex}{\large\bfseries\raggedright}}
\renewcommand{\subsection}{\@startsection{subsection}{2}{\z@}%
  {-1.8ex plus -0.5ex minus -0.2ex}{0.8ex plus 0.2ex}{\normalsize\bfseries\raggedright}}
\renewcommand{\subsubsection}{\@startsection{subsubsection}{3}{\z@}%
  {-1.5ex plus -0.5ex minus -0.2ex}{0.5ex plus 0.2ex}{\normalsize\bfseries\raggedright}}
\renewcommand{\paragraph}{\@startsection{paragraph}{4}{\z@}%
  {0.8ex plus 0.3ex minus 0.2ex}{-1em}{\normalsize\bfseries}}

\renewcommand{\maketitle}{%
  \begingroup
  \thispagestyle{plain}%
  \vbox{%
    \hsize\textwidth\linewidth\hsize
    \vskip 0.1in
    \hrule height 4pt
    \vskip 0.25in
    \centering{\LARGE\bfseries\@title\par}%
    \vskip 0.29in
    \hrule height 1pt
    \vskip 0.3in
    {\centering\@author\par}%
    \vskip 0.3in}%
  \endgroup}

\renewenvironment{abstract}{%
  \centerline{\large\bfseries Abstract}\vspace{0.5ex}%
  \begin{list}{}{\leftmargin=0.5in\rightmargin=0.5in\listparindent=0pt\parsep=5.5pt}\item\relax}%
  {\end{list}\vspace{1ex}}
\makeatother

\title{Distillation, Self-Training, and the Fragility of Learned Chemical Reasoning}

\author{%
  \begin{tabular}{c}
  \textbf{Anagha Aneesh}\quad
  \textbf{Sebastian Ibarraran}\quad
  \textbf{Frank Hu}\quad
  \textbf{Jason Liu}\quad
  \textbf{Grant M.\ Rotskoff}$^{*}$\\[2pt]
  Stanford University
  \end{tabular}}

\begin{document}

\maketitle
{\renewcommand{\thefootnote}{\fnsymbol{footnote}}%
\footnotetext[1]{Corresponding author: Grant M.\ Rotskoff, \href{mailto:rotskoff@stanford.edu}{\texttt{rotskoff@stanford.edu}}.}}

\begin{abstract}
Automating synthesis planning for small-molecule drug design requires chemical reasoners that are accurate as well as open to inspection and adaptation, yet frontier language models remain unreliable at predicting reaction outcomes and difficult to deploy, while small open-weight models lack the necessary domain reasoning. We show that frontier supervision can serve as a one-time bootstrap for open models on forward reaction prediction. Our models, \textbf{Resonance}, distill an open-weight teacher's reasoning traces and then learn from their own verified outputs, turning a small base model with near-zero accuracy into a capable reaction predictor, while a stronger student surpasses its teacher through distillation alone. Accuracy, however, does not bring robust reasoning: the small student inherits its teacher's single-pass style, self-training mainly sharpens answers it could already reach, and it readily abandons correct answers under pushback, suggesting that chemical reasoning models should be evaluated on the robustness of their reasoning as well as on the accuracy of their answers.
\end{abstract}

\section{Introduction}
Chemical foundation models are now widely used for molecular discovery tasks, but designing small organic molecules for pharmaceutical applications often requires iterative and time-consuming chemical synthesis planning, which these models do not provide. While frontier large language models (LLMs) have acquired substantial chemical knowledge and reasoning ability, they are far from reliable for predicting chemical synthesis \citep{li2025beyond}. Furthermore, their closed weights and increasing restrictions in the sciences make them difficult to adapt and deploy as components of specialized workflows. Smaller open-weight models avoid these constraints but perform poorly on domain-specific reasoning without further training \citep{yang2024qwen25math}. There is an urgent need for robust chemical reasoners to realize automated scientific applications.

Forward reaction prediction, predicting the major product of a chemical reaction from its reactants, is a core problem in computer-aided synthesis planning \citep{coley2018machine}. Progress on this problem supports retrosynthesis \citep{gao2024generative, genheden2020aizynthfinder}, route scoring \citep{segler2018planning, bran2026chemical}, and autonomous experimentation \citep{boiko2023autonomous}. The task is also a natural probe of chemical reasoning in language models because a model must recognize which functional groups react, infer the most likely transformation, and choose between competing products. Specialist sequence-to-sequence models trained on reaction SMILES already predict reaction products with high accuracy \citep{schwaller2019molecular}, but they provide answers without explanation and cannot be instructed or questioned. Models that reason in natural language pair each prediction with a rationale that can be inspected and composed into larger chemical workflows \citep{bran2026chemical, boiko2023autonomous}.

\begin{figure}[t]
    \centering
    \includegraphics[width=\linewidth]{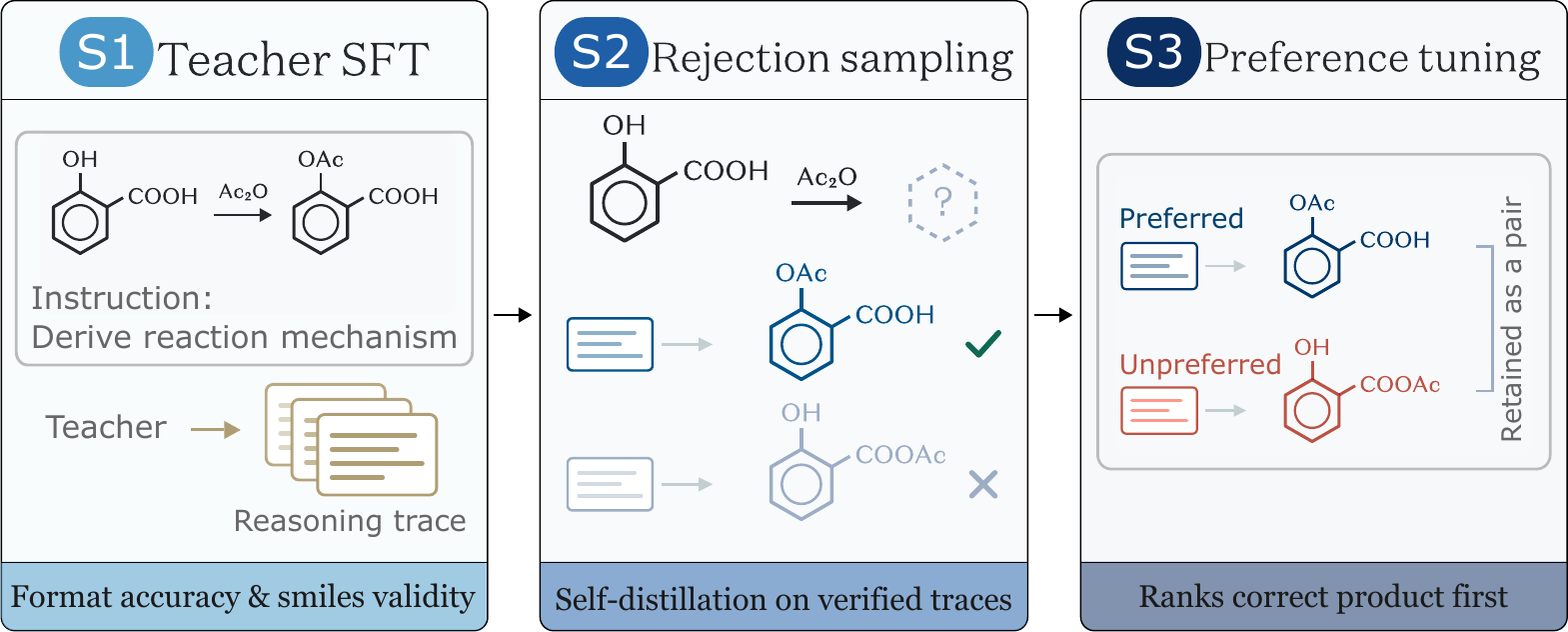}
    \caption{\textbf{The pipeline used to train Resonance.} The teacher (Kimi K3) generates reasoning traces, provided with the recorded product visible, that are used to fine-tune the base model (base $\rightarrow$ S1). The model is then improved on its own verified-correct traces (S1 $\rightarrow$ S2, rejection-sampling SFT) and its own preference-ranked traces (S2 $\rightarrow$ S3, DPO) before evaluation.}
    \label{fig:workflow}
\end{figure}

One way to close this gap is to distill a frontier model's reasoning traces into a smaller model \citep{hsieh2023distilling} and let it continue learning from its own verified outputs \citep{zelikman2022star, huang2024selfimprovement}. Final accuracy establishes whether this works, but reveals little about why. It does not show which post-training stage is responsible for improvements, whether the answer depends on the model being trained, or how the resulting model reasons. We study these questions by training \textbf{Resonance}, a family of open-weight reasoning models for forward reaction prediction (Figure~\ref{fig:workflow}). A teacher, Kimi K3 \citep{kimi2026k3}, supplies reasoning traces for supervised fine-tuning of a student (Stage S1), which then trains on its own verified-correct outputs through rejection-sampling fine-tuning (Stage S2) and preference optimization (Stage S3). We train Resonance-8B from Qwen3-8B \citep{qwen3technicalreport} and, with identical teacher traces, Resonance-27B from the stronger Qwen3.8-27B \citep{qwen38}.

Our contributions are:
\begin{itemize}[itemsep=3pt, topsep=3pt, parsep=0pt, partopsep=0pt]
  \item \textbf{Resonance}, open-weight reasoning models for forward reaction prediction and a protocol to build them by distilling a frontier teacher's reasoning traces and then self-training on the student's own verified outputs.
  \item \textbf{A stage-by-stage analysis of post-training}, showing that self-training substantially improves a weak student but adds little to a stronger one, and that the difference can be anticipated from how much of the training set the student already solves.
  \item \textbf{An analysis of how the student reasons}, showing that it inherits the reasoning style of its training traces and that its accuracy gains do not introduce robustness to prompting or feedback.
\end{itemize}

\section{Related Work}
\textbf{Reaction prediction and LLMs for chemistry.} Forward reaction prediction is now commonly framed as translation between reactant and product SMILES \citep{weininger1988smiles}, an approach popularized by the Molecular Transformer \citep{schwaller2019molecular}, which outperformed earlier template-based and template-free approaches \citep{liu2017retrosynthetic, segler2017neural}. Chemical language models pre-trained on large molecular corpora learn transferable molecular representations \citep{ross2022large}, and general-purpose LLMs have since been applied to reaction prediction and other chemistry tasks \citep{jablonka2024leveraging, guo2023can, wang2025survey, hu2026training}. Prior work adapts general models to chemistry through continued pretraining on scientific corpora \citep{taylor2022galactica}, instruction tuning on curated chemistry datasets \citep{zhang2025chemllm,bran2025mist}, or reinforcement learning on verifiable chemistry tasks \citep{narayanan2025training}. We instead start from general-purpose models and add domain competence only through post-training on the target task.

\textbf{Distillation, self-distillation and preference optimization.} Distilling a teacher's reasoning traces into a smaller student is a standard training recipe \citep{hinton2015distilling, hsieh2023distilling} and underpins recent open reasoning models \citep{guo2025deepseek}. Students can also improve on their own verified outputs through rejection sampling fine-tuning \citep{zelikman2022star, singh2023beyond} or iterative self-evolution on verified trajectories \citep{guan2025rstar}. Recent work frames such self-improvement as sharpening the model's distribution toward answers it already produces \citep{huang2024selfimprovement, yue2026does}. Direct Preference Optimization (DPO) \citep{rafailov2024direct} replaces the reward model with pairwise preferences and has been applied to self-generated preferences \citep{yuan2024selfrewarding, wu2025selfplay} and to pairs of correct and incorrect reasoning chains \citep{pang2024iterative}. We combine these steps in a single pipeline and measure what each contributes.

\textbf{Reasoning intervention.} Chain of thought (CoT) explanations are not always faithful to what drives a model's answer \citep{turpin2023language, chen2025reasoningmodelsdontsay}, and reasoning models have less control over their traces than over their final outputs \citep{yuehhan2026reasoningmodelsstrugglecontrol}. Test-time strategies help most when a model has been trained to use them; specifically, fine-tuning on long reasoning traces allows a model to be prompted to extend its reasoning and improve its accuracy as a result \citep{muennighoff2025s10}. Model improvement through self-training depends on whether it already verifies and backtracks \citep{gandhi2025cognitive}. Without such training, models struggle to correct their own reasoning \citep{huang2023large} and tend to change their answers when challenged \citep{sharma2024towards}. We study these interventions in a single scientific domain and across successive post-training stages, asking how distillation and self-training shape the way the resulting models reason and respond to challenge.

\section{Methods}
\label{sec:methods}

\subsection{Task formulation}
\label{sec:task}
We study forward reaction prediction: given reactant and reagent molecules encoded as SMILES \citep{weininger1988smiles}, predict the SMILES of the major product. We use USPTO-50k with its standard split of 40{,}000 training, 5{,}000 validation and 5{,}000 test reactions. Each reaction's reactants are split into a starting material (the first reactant) and reagents (all other reactants). This version of the dataset records no solvents, catalysts or conditions, so when a reaction has a single reactant the prompt reads ``Reagents: No reagents'' (Appendix~\ref{app:data}).

We canonicalize all molecules with RDKit~\citep{rdkit} and score a prediction as correct only if its canonical SMILES exactly matches the canonical ground truth reaction product (Top-1 exact match). Over $N$ test reactions
\begin{equation*}
  \mathrm{Acc@1} = \frac{1}{N}\sum_{i=1}^{N} \mathbbm{1}\left[\mathrm{canon}(\hat{y}_i) = \mathrm{canon}(y_i)\right],
\end{equation*}
with $\hat{y}_i$ the predicted product, $y_i$ the ground truth, and $\mathrm{canon}(\cdot)$ is the RDKit canonicalization function. The prediction is the last \texttt{<answer>} block; invalid SMILES, missing answers and truncated generations are scored as incorrect and stay in the denominator.

\subsection{Base models and teacher}
We start from an open-weight Qwen model, either Qwen3-8B or Qwen3.8-27B, and elicit predictions in a CoT~\citep{wei2022chainofthought} analysis followed by a delimited final SMILES answer that the scorer extracts. The two models differ in pre-training recipe as well as size, so we treat them as two starting points rather than as a scaling series. We use Kimi K3 as the teacher because it is a strong open-weight model whose traces we can release in full, so the supervision used in Stage~1 is available to anyone reproducing this work. Kimi K3, Claude Opus 5~\citep{anthropic2026claudeopus5} and GPT-6-Astra~\citep{openai2026gpt6astra} are also evaluated zero-shot on the test set with the product hidden (Appendix~\ref{app:eval}). Training details can be found in Appendix~\ref{app:train}.

\subsection{Three stages of training}
\label{sec:stages-method}
\textbf{Stage 1: SFT on teacher traces.}
We generate reasoning traces for all 40{,}000 training reactions with Kimi K3, showing it the starting material, reagents and recorded product (prompt in Appendix~\ref{app:prompts}), and fine-tune the Qwen model on each trace followed by the recorded product, yielding S1. This installs the trace format and an initial competence the small pre-trained model lacks, effectively moving the model on-distribution for the reaction prediction task.

\textbf{Stage 2: Rejection-sampling SFT on self-generated traces.}
We sample 8 traces per training reaction from S1, keep only those whose product is verified as correct, and fine-tune on the retained set to yield S2. This stage reinforces reasoning paths that actually reach correct products and shifts the training data on-policy. For the 8B model it keeps 47{,}492 examples covering 19{,}668 reactions. Because the 27B model is already correct on 85.8\% of its S1 samples, naive rejection sampling would mostly retrain it on solved reactions. We therefore re-balance its data toward reactions it solves in some, but not all, of 8 traces and provide the ground-truth product in an answer-only format for the 1{,}108 reactions with no correct sample (1.1\% of its S2 examples; Appendix~\ref{app:s2-27b}).

\textbf{Stage 3: DPO on self-generated preference pairs.}
We sample 16 traces per training reaction from S2. For each reaction with at least one correct and one incorrect sample, we take a random verified-correct trace as the chosen response $y_w$ and the incorrect trace whose product is most similar to the recorded product as the rejected response $y_l$, keeping one pair per reaction. We optimize the DPO objective \citep{rafailov2024direct} together with a negative log-likelihood term on the chosen response:
\begin{equation*}
  \mathcal{L}_{\mathrm{S3}} = -\,\mathbb{E}_{(x,\,y_w,\,y_l)}\!\left[ \log \sigma\!\left( \beta \log \frac{\pi_\theta(y_w \mid x)}{\pi_{\mathrm{ref}}(y_w \mid x)} - \beta \log \frac{\pi_\theta(y_l \mid x)}{\pi_{\mathrm{ref}}(y_l \mid x)} \right) + \lambda\, \frac{1}{|y_w|} \log \pi_\theta(y_w \mid x) \right],
\end{equation*}
where $\pi_{\mathrm{ref}}$ is the S2 checkpoint, $\beta = 0.1$ controls regularization to the reference policy, and $\lambda = 0.2$ weights the per-token log-likelihood of the chosen response. The preference term uses log-probabilities summed over the response; the likelihood term, similar to~\cite{pang2024iterative}, keeps the probability of correct traces from falling during preference optimization.

\subsection{Reasoning analysis}
\label{sec:reasoning-methods}
We probe Resonance-8B at each stage with prompts that change how it reasons, feedback on its answers, and repeated sampling. All probes use the evaluation prompt and scorer, temperature $0.6$, Top-$p=0.95$, Top-$k=20$, an 8{,}192-token output cap, and an independent seed for each reaction, with 8 draws per reaction unless stated otherwise (Appendix~\ref{app:eval}). Each condition is compared with a baseline on the same reactions, using paired tests clustered by reaction (Appendix~\ref{app:stats}). Probe baselines differ slightly from the single-draw headline accuracies (Appendix~\ref{app:eval}).

\textbf{Reasoning-length prompts.}
We append an instruction asking the model to \textbf{commit} to its first well-supported answer, to \textbf{consider alternatives} before finalizing, or to \textbf{think more} before committing. These instructions provide no chemical information. Each is compared with an unprompted resample of the same model.

\textbf{Feedback.}
After the model's first answer, we send a second user turn, regardless of that answer's correctness: \textbf{mild doubt} (``there may be an issue with it''), \textbf{told wrong}, \textbf{marked wrong} by an evaluator, or \textbf{told correct} (Appendix~\ref{app:prompts}). Each is compared with a neutral ``review your answer'' turn.

\textbf{Backtracking.}
We count a fixed list of backtracking phrases (for example ``wait'', ``let me reconsider'', ``alternatively'') inside the thinking block, frozen before any intervention was run, and report events per 1{,}000 output tokens and the share of traces with at least one event (Appendix~\ref{app:lexicon}). This is a surface measure that we use only to compare conditions.

\textbf{Repeated sampling.}
We draw nine predictions per reaction (the first answer plus 8 resamples) and evaluate plurality voting for $k \in \{1,3,5,7,9\}$ (\textbf{plurality@$k$}): the most frequent canonicalized product among the first $k$ samples is selected, excluding unparsable outputs and resolving ties by earliest occurrence. We also report \textbf{pass@$k$}, the fraction of reactions with at least one correct prediction among the first $k$ samples.

\section{Results}
\label{sec:results}

\begin{figure}[t]
    \centering
    \includegraphics[width=0.8\linewidth]{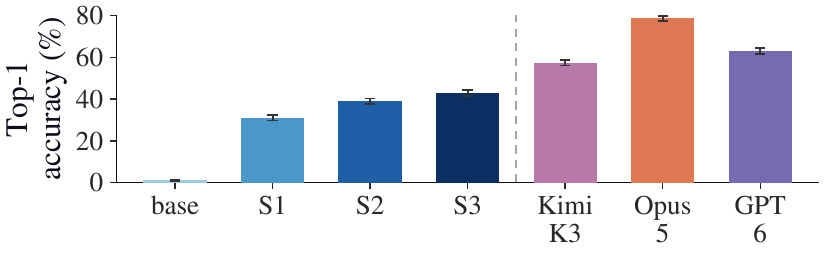}
    \caption{\textbf{Top-1 exact-match accuracy on the USPTO-50k test set} ($N=5{,}000$) for Resonance-8B at each stage (base $\rightarrow$ S1 teacher SFT $\rightarrow$ S2 rejection-sampling SFT $\rightarrow$ S3 DPO), and for Kimi K3 (teacher, 57.4\%), Claude Opus 5 (78.7\%) and GPT-6-Astra (63.0\%), all three evaluated zero-shot. Each stage improves on the previous one, and the two self-training stages together add 11.8 points without teacher supervision. Error bars are 95\% Wilson intervals.}
    \label{fig:accuracy}
\end{figure}

\subsection{Self-training improves the 8B model after teacher-guided SFT}
\label{sec:stages}

We evaluate on the 5{,}000-reaction test split of USPTO-50k, a standard benchmark of roughly 50{,}000 reactions drawn from US patents \citep{lowe2012extraction, schneider2016whats}. A prediction is evaluated as correct if its RDKit-canonical SMILES exactly matches the recorded product. We compare Resonance-8B at each stage with its teacher, Kimi K3, and with two frontier models, Claude Opus 5 \citep{anthropic2026claudeopus5} and GPT-6-Astra~\citep{openai2026gpt6astra}.

Out of the box, Qwen3-8B is unable to complete this forward reaction prediction task. It predicts 1.2\% of products correctly (Figure~\ref{fig:accuracy}). Few-shot examples raise its format compliance to 89\% of finished generations but leave accuracy at 1--2\%, and over half of its valid answers simply copy the starting material (Appendix~\ref{app:fewshot}). Teacher SFT (S1) changes this, lifting accuracy to 31.0\%. For this small model, S1 is an essential step in establishing baseline task performance that supports further gains via self-training. Rejection sampling (S2) and preference optimization (S3) can only learn from the model's self-produced correct answers, and with the base model alone (1.2\% accuracy) there is almost nothing to learn from. After S1, the model is able to learn from its own verified work. S2 adds 7.9 points and S3 another 3.9, reaching 42.8\% accuracy overall, and each stage is a significant improvement on the one before (Appendix~\ref{app:stage-tests}). More than a quarter of the total improvement (11.8 of 41.6 points) arrives after the teacher has left the training loop. Rejection sampling is needed: applying DPO directly to S1, without S2, recovers only about half of this gain (Appendix~\ref{app:skip-s2}). Despite these gains, the performance of the student model still lags behind that of the teacher model. This raises the question of whether the ceiling of student model performance is set by its pretraining recipe or if this is an artifact of the teacher model traces. Self-training alone cannot answer this question because it only reinforces reaction knowledge the S1 model can already reach (Section~\ref{sec:reasoning}). We return to this question in Section~\ref{sec:27b}, where the same traces are used to train a stronger model.

\begin{figure}[t]
    \centering
    \includegraphics[width=\linewidth]{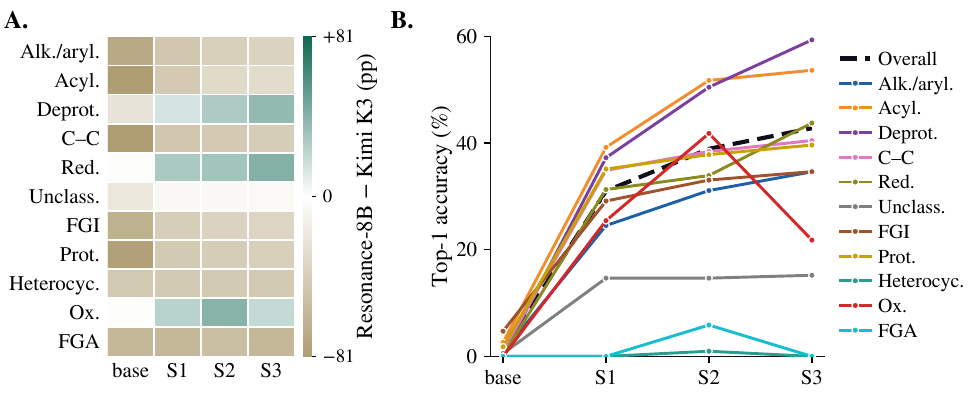}
    \caption{\textbf{Accuracy by reaction class for Resonance-8B.} (A) Student minus teacher zero-shot accuracy (Resonance-8B minus Kimi K3) at each stage; green marks a student advantage and tan a teacher advantage. (B) Student Top-1 accuracy by reaction class across training stages. Classes follow Table~\ref{tab:classes}, which also gives training and test counts; all 5{,}000 test reactions are included.}
    \label{fig:reaction_data}
\end{figure}

\subsection{Students and teacher understand different chemistry}
\label{sec:families}
Aggregate accuracy hides how unevenly the student's gains are distributed across reaction classes (see Figure~\ref{fig:reaction_data} and Table~\ref{tab:per-class}). Resonance-8B trails Kimi K3 on most classes but overtakes it, by wide margins, on deprotections, reductions, and oxidations. These three classes expose a limitation of USPTO-50k: the dataset often omits the reagent that drives the transformation, and the reducing, oxidizing, or deprotecting agent is missing from most of these prompts. Given such a prompt zero-shot, Kimi K3 typically concludes that no reaction occurs, which is a genuine error rather than a scoring artifact (Appendix~\ref{app:teacher-errors}). Yet, when generating training traces with the product provided, the same teacher can aptly explain the transformation and transfer knowledge to the student. The classes also respond differently to self-training (Figure~\ref{fig:reaction_data}B; Table~\ref{tab:per-class}). For most classes, the gain comes at S2: deprotections improve the most overall (22.0 percentage points from S1 to S3), and acylations make almost all of their 14.5-point gain at S2. Reductions are the one large class that improves mainly at S3, gaining 2.6 points at S2 and 9.9 at S3. Oxidations go the other way, gaining 16.3 points at S2 and then losing 20.0 at S3 to end below S1.

\subsection{Naive test-time scaling after post-training}
\label{sec:reasoning}
A common way to improve a reasoning model without further training is to spend more compute at inference time: asking it to think more, to consider alternatives, or to revise its answer in response to feedback. We test these strategies naively, through prompting alone, on a model that was never trained to use them (Section~\ref{sec:reasoning-methods}). Most do not improve Resonance-8B's accuracy at any stage and incur a slight performance cost. The reason appears to lie in the reasoning style the student inherits from its training data.

\paragraph{The student inherits a single-pass reasoning style.}
We measure how often a trace reconsiders its own reasoning by counting backtracking phrases such as ``wait'' or ``let me reconsider'' (Section~\ref{sec:reasoning-methods}). Qwen3-8B backtracks in every trace, but the Kimi K3 traces it is fine-tuned on rarely do: only 11\% contain a backtracking phrase (Table~\ref{tab:style}). After S1, the student matches its training data rather than its base model, backtracking in 13\% of traces, and self-training barely changes this (15\% at S3). The teacher's traces may lack backtracking because we asked for concise, fixed-format traces, or because the teacher could see the product while writing them.

\begin{figure}[t]
    \centering
    \includegraphics[width=\linewidth]{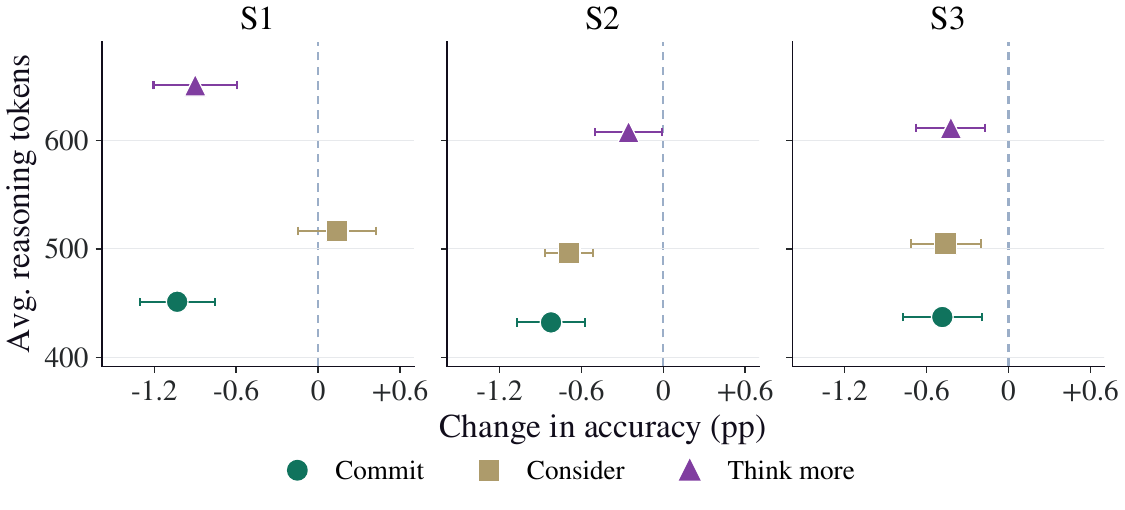}
    \caption{\textbf{Instructions change reasoning length, not accuracy.} Average reasoning tokens against change in Top-1 accuracy relative to an unprompted resample, for Resonance-8B at each stage, when instructed to \textbf{commit} to its first supported answer, \textbf{consider} alternatives, or \textbf{think more}. Baseline accuracy is 29.0\%, 38.8\% and 41.3\% at S1, S2 and S3, respectively. No instruction improves accuracy at any stage. Error bars are $\pm 1$ standard error over 8 draws per reaction, clustered by reaction.}
    \label{fig:prompting}
\end{figure}

\paragraph{Longer reasoning is not better reasoning.}
Asking the model to think more produces the longest traces at every stage (651 average tokens at S1, against 488 for an unprompted resample), asking it to commit produces the shortest (451 tokens), and asking it to consider alternatives falls in between (Figure~\ref{fig:prompting}). None of these instructions improves accuracy at any stage, and most lower it by up to about one percentage point (Table~\ref{tab:length-prompts}), suggesting that pushing the model away from the way it was trained to reason slightly reduces accuracy. Asking the model to consider alternatives raises its backtracking about 2.4-fold (0.27 to 0.66 events per 1{,}000 tokens at S1) with no gain in accuracy (Figure~\ref{fig:a2}). The student can mimic reconsidering when asked, but not constructively.

\begin{figure}[b]
    \centering
    \includegraphics[width=\linewidth]{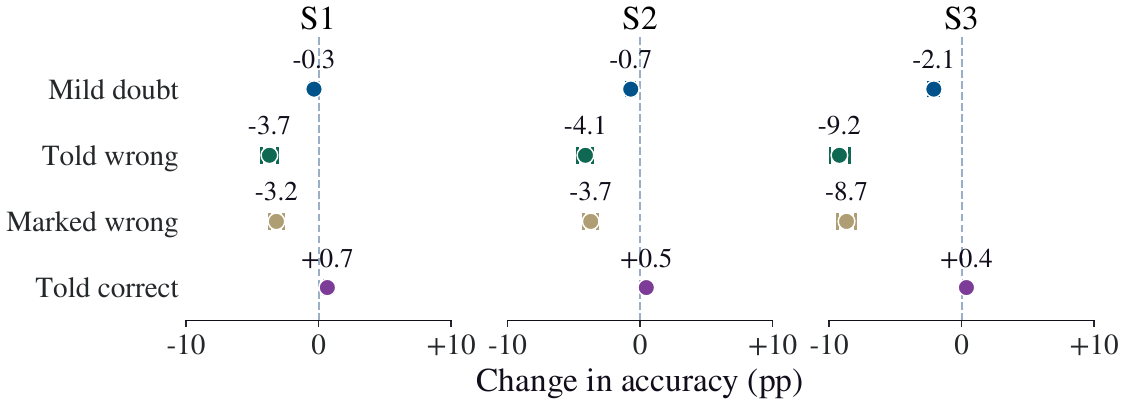}
    \caption{\textbf{Sensitivity to negative feedback grows with training.} Change in Top-1 accuracy of Resonance-8B relative to a neutral ``review your answer'' turn when its previous answer is returned with mild doubt, a statement that it is wrong, a statement that an evaluator marked it wrong, or a statement that it is correct, regardless of its actual correctness. Baseline accuracy after the neutral review is 28.0\%, 38.3\% and 41.1\% at S1, S2 and S3, respectively, with 8 draws per reaction. Negative feedback costs 3.2--3.7 points at S1 and 8.7--9.2 at S3; being told an answer is correct is the only intervention that helps, by less than one point. Error bars are $\pm 1$ standard error over 8 draws per reaction, clustered by reaction.}
    \label{fig:feedback}
\end{figure}

\paragraph{Pushback, not evidence, changes the model's answer.}
We next return the model's previous answer to it with feedback attached, regardless of whether that answer was correct (Figure~\ref{fig:feedback}). Because the feedback carries no information about the reaction, an ideal model would ignore it. Being told an answer is correct is the only intervention that asks the model to behave according to its training and is the only one that improves performance. Negative feedback, however, is costly. When told that its answer is wrong, the model loses 3.7 points at S1, 4.1 at S2 and 9.2 at S3; told that an evaluator marked it wrong, it loses 3.2, 3.7 and 8.7 (Table~\ref{tab:feedback}). Even a mild suggestion that the answer may contain an error costs more as training proceeds, rising from 0.3 to 0.7 to 2.1 points. The loss is not simply a consequence of S3 having more correct answers to lose: as a percentage of baseline accuracy, the \textbf{told wrong} loss is 13\% at S1 and 11\% at S2 but 22\% at S3. Separating the two directions of change shows what the model does with the feedback (Table~\ref{tab:flips}). Told that it is wrong, the model abandons 22--32\% of its initially correct answers, against 3--6\% after a neutral review, yet rescues only about 6\% of its initially incorrect ones. It treats negative feedback as an instruction to change its answer rather than as evidence to weigh, a form of sycophancy \citep{sharma2024towards}, and training does not remove this behavior at any stage.

\begin{figure}[t]
    \centering
    \includegraphics[width=\linewidth]{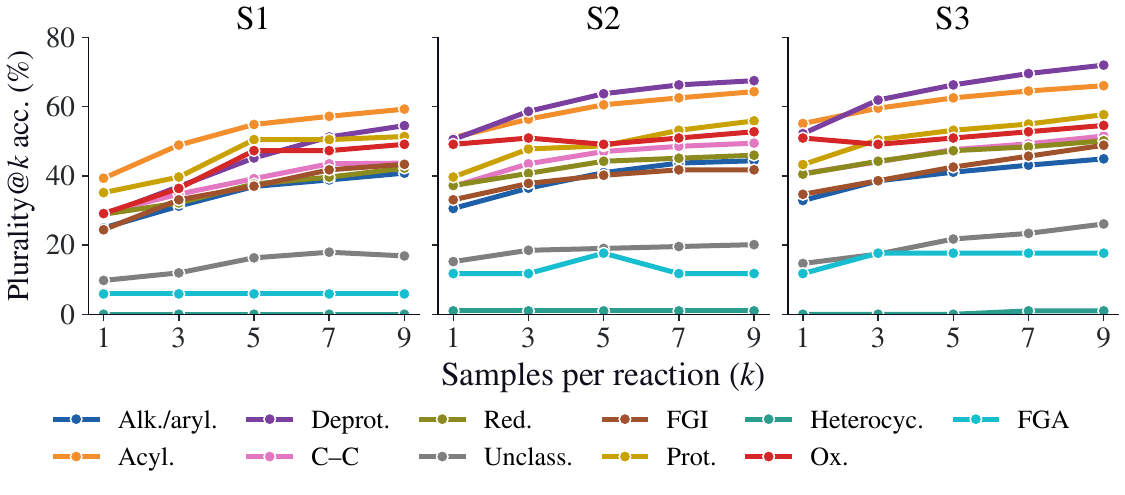}
    \caption{\textbf{Self-training sharpens answers the model could already reach.} Plurality@$k$ accuracy (the most frequent product among $k$ samples) for Resonance-8B by reaction class and stage. Voting over nine S1 samples matches or exceeds single-sample S3 accuracy in 8 of the 10 classes with at least 40 test reactions; heterocycle formation remains unsolved at every $k$. Classes follow Table~\ref{tab:classes}.}
    \label{fig:plurality}
\end{figure}

\paragraph{Self-training sharpens what the model could already find.}
Finally, we sample each reaction nine times to ask whether self-training teaches the model new reactions or makes it more reliable on reactions it could already solve (Figure~\ref{fig:plurality}; Appendix~\ref{app:sampling}). Between S1 and S3, single-sample accuracy rises by about 13 percentage points, but the share of reactions solved in at least one of nine samples rises by only about 4. In most reaction classes, a plurality vote over nine S1 samples already matches or exceeds a single S3 sample. Self-training therefore sharpens the model's distribution toward answers it can already reach rather than teaching it new ones \citep{huang2024selfimprovement}, which is why its gains in Section~\ref{sec:stages} are real but bounded. Heterocycle formation, where the 8B model almost never samples a correct product, shows this limit most clearly. The sharpening is also incomplete because plurality voting still improves S3 accuracy substantially, and this gain is available at inference time without further training.

\subsection{A stronger starting point needs less of the pipeline}
\label{sec:27b}

We apply the same pipeline, with the same teacher traces, to Qwen3.8-27B. Teacher SFT alone takes this model from 26.3\% to 79.0\%, well above Kimi K3, whose traces it was trained on (Figure~\ref{fig:27b}; stage-level accuracies are in Appendix Figure~\ref{fig:27b-acc}), which answers the question left open in Section~\ref{sec:stages}: the teacher's traces contain more than the 8B model learns from them, so the 8B model's ceiling is set by pre-training rather than by its supervision. The same holds within reaction classes, since heterocycle formation, which the 8B model never learns, reaches 42.7\% on Resonance-27B after S1 on identical traces, and part of the 27B student's margin over its teacher comes from the traces themselves rather than from the recorded products alone (Section~\ref{sec:families}). After S1 the model is also comparable to Claude Opus 5 (78.7\%), although the student is trained on USPTO-50k while Opus 5 is evaluated zero-shot.

\begin{figure}[t]
    \centering
    \includegraphics[width=\linewidth]{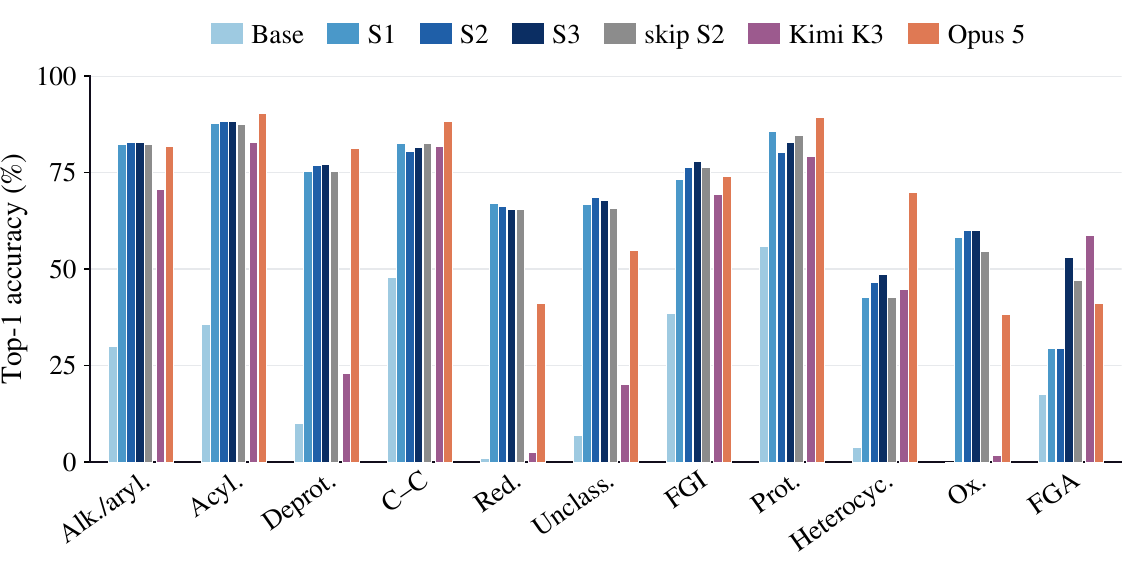}
    \caption{\textbf{Accuracy by reaction class for Resonance-27B.} Bars show the base model, S1, S2, S3, DPO applied directly to S1 (skip-S2); Kimi K3 and Claude Opus 5 (zero-shot). Teacher SFT supplies nearly all of the gain in every class, and the later stages change accuracy by amounts within sampling noise. Opus 5 leads on heterocycle formation, while the student leads on reductions and oxidations, where the prompt usually names no reagent. Classes follow Table~\ref{tab:classes}.}
    \label{fig:27b}
\end{figure}

Neither self-training stage adds measurable accuracy to Resonance-27B: S2 reaches 79.3\% and S3 79.6\%, differences too small to distinguish from sampling noise, and skipping S2 gives the same result (Appendices~\ref{app:skip-s2} and~\ref{app:stage-tests}). Heterocycle formation rises from 42.7\% to 48.5\% across these stages, but with only 103 test reactions this change is also within noise. Self-training has little to learn from once the 27B model has been distilled, since after S1 it already answers most of its training reactions correctly (85.8\% of samples, against 16.6\% for the 8B model; Table~\ref{tab:s1-solve}) and few reactions yield the mix of correct and incorrect samples that rejection sampling and preference optimization need. We rebalanced its S2 data toward reactions it solves only some of the time to give self-training its best chance, choosing the recipe after several variants (Appendix~\ref{app:s2-27b}), but it still adds no measurable accuracy. What S1 does provide the 27B model, in part, is concision: before training, fewer than half of its generations reach an answer within the token budget, whereas after fine-tuning on the teacher's short, single-pass traces it answers within a few hundred tokens. For this model, the reasoning style that limits the 8B student's robustness (Section~\ref{sec:reasoning}) may therefore be part of what makes it effective.

\FloatBarrier

\section{Conclusions}
\label{sec:conclusions}
Our distillation and self-training pipeline transforms a small open-weight model that cannot perform forward reaction prediction into a capable one using a modest, task-specific dataset. This approach matters as agentic methods increasingly rely on systems of many small models \citep{belcak2025small}, which in scientific domains must often be specialized with far less data than general-purpose pre-training provides~\citep{vanherck2025assessment}. How much each stage contributes, however, depends on the student. Qwen3-8B needs teacher SFT to begin the task at all, and self-training then adds substantially more without further supervision, whereas Qwen3.8-27B surpasses its teacher after SFT alone and gains nothing measurable from the later stages. Because self-training helps only while the distilled model still fails on much of its training set, and that failure rate can be measured after S1, it offers a practical guide to whether further post-training is worth the cost. Benchmark accuracy, however, does not show how these models reason. The 8B student inherits the single-pass style of its training traces, so prompts to reason longer or consider alternatives do not help, and self-training mainly sharpens answers the model could already reach by plurality vote. When told a correct answer is wrong, the student abandons 22--32\% of such answers at every stage, and the accuracy lost under pushback more than doubles between S1 and S3. We hypothesize that the missing capability is meta-reasoning, knowing when and how to revise: the training corpus rarely demonstrates revision (Appendix~\ref{app:style}), prompted deliberation raises the surface markers of reconsidering without raising accuracy (Appendices~\ref{app:length-prompts} and~\ref{app:revisions}), and negative feedback flips answers rather than prompting a check (Appendix~\ref{app:feedback}). These observations are consistent with the hypothesis but do not establish it, and they suggest that models intended to reason about chemistry should be evaluated on the robustness of their reasoning as well as on whether they reach the correct product.

\paragraph{Limitations.}
The students are trained on USPTO-50k while the frontier models are evaluated zero-shot, so comparisons with them measure what task-specific post-training adds rather than general chemical ability. USPTO-50k often omits the reagent that drives a reaction, so part of what a trained model learns is how the dataset records such reactions, and zero-shot models are penalized for declining to infer a missing reagent. We evaluate a single task and benchmark, train each configuration once, and score only the final product without assessing reasoning traces for accuracy beyond the ground truth. Finally, our reasoning analysis covers only the 8B model, and its backtracking measure is a lexical count not yet validated against human labels.

\paragraph{Future work.}
Testing the meta-reasoning hypothesis requires training data that demonstrates revision directly, such as failed synthetic pathways, a diagnosis of why each failed, and recovery from them; a strong benchmark score does not certify a teacher as a source of such data. Repeating the reasoning analysis on the 27B model would show whether a stronger student responds differently to feedback, and evaluating on reaction data that records full conditions would separate chemical reasoning from learning dataset conventions. For capable students that already solve most of their training set, self-training will need new signal, for example from harder or out-of-distribution chemistry, and scoring intermediate mechanistic steps rather than only the final product would bring evaluation closer to the reasoning these models are meant to perform.

\section*{Acknowledgments}
This work was supported by a Stanford HAI grant and by NIH grant 5R35GM159834. A.A.\ and S.I.\ acknowledge support from the NSF Graduate Research Fellowship. We also acknowledge a \$10k Academic Grant provided by Modal.

\clearpage

\bibliographystyle{plainnat}
\bibliography{references}

\appendix
\newtcolorbox{qbox}{enhanced,breakable,colback=gray!5,colframe=gray!55,
  boxrule=0.4pt,arc=1.5pt,left=5pt,right=5pt,top=3pt,bottom=3pt,
  fontupper=\small}
\providecommand{\qhead}[3]{%
  \begin{qbox}
  \textbf{Question.} #1\par\smallskip
  \textbf{Supports.} #2\par\smallskip
  \textbf{Takeaway.} #3
  \end{qbox}}

\lstdefinestyle{prompt}{basicstyle=\ttfamily\scriptsize,breaklines=true,
  frame=single,rulecolor=\color{gray!55},columns=fullflexible,
  keepspaces=true,xleftmargin=2pt,xrightmargin=2pt}
\lstdefinestyle{trace}{style=prompt,
  literate={–}{{\textendash}}1 {—}{{\textemdash}}1 {₂}{{$_2$}}1}
\clearpage
\section*{Appendix}
\addcontentsline{toc}{section}{Appendix}

\paragraph{Guide to the appendix.}
Each section answers one question a reader may have about the main
results. Table~\ref{tab:guide} maps questions to sections and to the
main-text claims they support.

\begin{table}[h]
\centering\small
\caption{\textbf{Guide to the appendix.}}
\label{tab:guide}
\begin{tabular}{p{0.52\linewidth}p{0.12\linewidth}p{0.22\linewidth}}
\toprule
Question & Section & Supports \\
\midrule
\multicolumn{3}{l}{\emph{Setup}}\\
How were the data scored and the reaction classes assigned? & \ref{app:data} & \S\ref{sec:task}, all per-class results\\
What exactly was trained, with what settings? & \ref{app:train}--\ref{app:eval} & \S\ref{sec:stages-method}, reproducibility\\
What prompts were used? & \ref{app:prompts} & \S\ref{sec:reasoning-methods}, \S\ref{sec:reasoning}\\
\midrule
\multicolumn{3}{l}{\emph{What does each training stage contribute?}}\\
Does the base model fail only because of output format? & \ref{app:fewshot} & \S\ref{sec:stages}\\
Is rejection-sampling SFT (S2) necessary? & \ref{app:skip-s2} & \S\ref{sec:stages}\\
Why does S2 differ for the 27B model? & \ref{app:s2-27b} & \S\ref{sec:27b}\\
Are the differences between stages significant? & \ref{app:stage-tests} & \S\ref{sec:stages}, \S\ref{sec:27b}\\
How does the 27B model's accuracy change by stage? & \ref{app:27b-acc} & \S\ref{sec:27b}\\
\midrule
\multicolumn{3}{l}{\emph{Where do the students beat the teacher?}}\\
Accuracy by class for every model and stage & \ref{app:per-class} & \S\ref{sec:families}\\
Are the teacher's low scores real errors or scoring artifacts? & \ref{app:teacher-errors} & \S\ref{sec:families}\\
Do students and teacher solve the same reactions? & \ref{app:overlap} & \S\ref{sec:families}\\
\midrule
\multicolumn{3}{l}{\emph{How does the student reason?}}\\
How is backtracking measured, and is the measure valid? & \ref{app:lexicon} & \S\ref{sec:reasoning}\\
Where does the student's reasoning style come from? & \ref{app:style} & \S\ref{sec:reasoning}\\
Do prompts to reason longer or shorter change accuracy? & \ref{app:length-prompts} & \S\ref{sec:reasoning}\\
How does the model respond to feedback? & \ref{app:feedback} & \S\ref{sec:reasoning}\\
Do structure changes within a trace reach the right product? & \ref{app:revisions} & \S\ref{sec:reasoning}, \S\ref{sec:conclusions}\\
What does sampling several answers per reaction add? & \ref{app:sampling} & \S\ref{sec:reasoning}\\
\midrule
How were the statistics computed? & \ref{app:stats} & all\\
\bottomrule
\end{tabular}
\end{table}

\section{Experimental setup}
\label{app:setup}

\subsection{Data, scoring and reaction classes}
\label{app:data}

\paragraph{Dataset.}
USPTO-50k \citep{schneider2016whats}, standard split:
40{,}000 train, 5{,}000 validation and 5{,}000 test reactions. All 40{,}000
training reactions are used for S1. Each reaction's reactants are split into
a \emph{starting material} (the first reactant) and \emph{reagents} (all other
reactants, joined with ``.''). This version of the dataset records no
solvents, catalysts or conditions, so the ``reagents'' are in fact
co-reactants; when a reaction has a single reactant the prompt reads
``Reagents: No reagents'' (1{,}439 of 5{,}000 test prompts, 28.8\%).
The source files contain no atom maps; the scorers strip any that appear.

\paragraph{Scoring.}
The prediction is the last \texttt{<answer>} block. Prediction and
reference are canonicalized with RDKit (isomeric SMILES, atom maps removed)
and a prediction is correct only on exact string match. The whole
dot-separated answer is canonicalized as one molecule, so fragment order does
not matter but every fragment, including counter-ions, must match; no gold
product in train or test contains more than one component. Invalid SMILES,
missing answer tags and truncated generations with no answer are scored as
incorrect and stay in the denominator ($N = 5{,}000$).
The last \texttt{<answer>} tag anywhere in the response counts, including one
inside a thinking block that never closed; a generation truncated before any
answer tag is incorrect. One scorer is used for every model.
\emph{Sensitivity to this rule.} A stricter rule that reads only text after a
closing \texttt{</think>} tag leaves every trained 27B checkpoint unchanged
(S1/S2/S3 79.0/79.3/79.6\%, skip-S2 78.8\%) and the 8B base
model unchanged (1.2\%), but lowers the 27B base model from 26.3\% to 22.8\% and
the 8B students from 31.0 / 38.9 / 42.8\% to 26.2 / 34.5 / 34.9\% at S1/S2/S3
(skip-S2: 37.1 to 31.5\%). These are not truncations: at most one headline
response per stage exceeds 8{,}192 tokens. The model states its answer and does
not close the thinking block, which S3 does in 18.5\% of responses (S1 16.4\%,
S2 10.5\%). Under the stricter rule S1$\to$S2 remains significant
($+8.3$ [6.9, 9.8], $p<10^{-28}$) and skip-S2 remains below the full pipeline
($-3.5$ [$-4.8$, $-2.1$], $p<10^{-6}$), but S2$\to$S3 does not
($+0.4$ [$-1.0$, 1.8], $p=0.58$).
\paragraph{Reaction classes.}
We use one labeling scheme throughout the paper
(Table~\ref{tab:classes}). Labels come from a rule-based classifier that
finds the formed bond from functional-group and ring changes between the
reactant union and the product, followed by a per-fragment maximum common
substructure search; its class distribution on USPTO-50k is within
about 1--3 points of the published Schneider et al.\ distribution. Reactions that match no rule are labeled unclassified (184 test, 1,684 training reactions). This is not a chemical class: most are single-reactant transformations the rules do not encode, including aryl methyl ether cleavage, N-debenzylation, and smaller groups such as thiourea formation and MOM removal. We then correct one systematic error on the test set: benzyl ethers \texttt{[OX2;H0]Cc1ccccc1}) had been treated as protecting groups, which moves 282 of 5{,}000 test reactions (most often deprotections $\to$  acylation, 148, and protections $\to$ reductions, 34) and 2{,}247 of 40{,}000 training reactions.

\begin{table}[h]
\centering\small
\caption{\textbf{Reaction classes in USPTO-50k.} This scheme is used in every per-class figure and table; figures use the abbreviations. FG = functional group. Test and training labels both include the benzyl-ether correction. Unclassified = no rule of the classifier applies (see text).}
\label{tab:classes}
\begin{tabular}{llrr}
\toprule
Class & Abbrev. & $n$ (test) & $n$ (train) \\
\midrule
Heteroatom alkylation/arylation & Alk./aryl. & 1{,}479 & 11{,}736 \\
Acylation                       & Acyl.      & 1{,}107 &   8{,}733 \\
Deprotections                   & Deprot.    &   824 &   6{,}494 \\
C--C bond formation             & C--C       &   536 &   4{,}281 \\
Reductions                      & Red.       &   457 &   3{,}542 \\
Unclassified                    & Unclass.   &   184 &   1{,}684 \\
FG interconversion              & FGI        &   127 &   1{,}075 \\
Protections                     & Prot.      &   111 &      977 \\
Heterocycle formation           & Heterocyc. &   103 &        885 \\
Oxidations                      & Ox.        &    55 &        445 \\
FG addition                     & FGA        &    17 &        148 \\
\midrule
Total                           & & 5{,}000 & 40{,}000 \\
\bottomrule
\end{tabular}
\end{table}
\subsection{Training configurations}
\label{app:train}

Both models start from open-weight Qwen checkpoints (Qwen3-8B,
Qwen3.8-27B). Table~\ref{tab:train-config} gives all three stages side
by side so the differences between the two models are visible. The
reason S2 differs for 27B is given in App.~\ref{app:s2-27b}.
All runs use bfloat16, gradient checkpointing and seed 42.
Both S1 runs compute the loss on all non-padding tokens, including the
prompt. S2 computes it on the full sequence for 8B and on the assistant
completion only for 27B; S3 masks the prompt.

\begin{table}[h]
\centering\scriptsize
\setlength{\tabcolsep}{3.5pt}
\caption{\textbf{Training configuration by model and stage.} Differences between the models are in \textbf{bold}.}
\label{tab:train-config}
\resizebox{\linewidth}{!}{%
\begin{tabular}{lcccccc}
\toprule
& \multicolumn{2}{c}{S1 (teacher SFT)} & \multicolumn{2}{c}{S2 (RS-SFT)} & \multicolumn{2}{c}{S3 (DPO)}\\
\cmidrule(lr){2-3}\cmidrule(lr){4-5}\cmidrule(lr){6-7}
Parameter & 8B & 27B & 8B & 27B & 8B & 27B\\
\midrule
Fine-tuning        & full & full (text) & \textbf{full} & \textbf{LoRA} & LoRA & LoRA\\
Epochs / steps     & 3 ep (3{,}750 st) & 3 ep (3{,}750 st) & \textbf{3 ep (2{,}229 st)} & \textbf{1 ep (3{,}086 st)} & \textbf{3{,}000 st (3.3 ep)} & \textbf{1 ep (378 st)}\\
Learning rate      & $2\!\times\!10^{-5}$ & $2\!\times\!10^{-5}$ & $\mathbf{1\!\times\!10^{-5}}$ & $\mathbf{2\!\times\!10^{-5}}$ & $2\!\times\!10^{-6}$ & $2\!\times\!10^{-6}$\\
Warmup steps       & 100 & 100 & 50 & 50 & 0 & 0\\
Effective batch    & 32 & 32 & \textbf{64} & \textbf{32} & 32 pairs & 32 pairs\\
Max sequence length& \textbf{2{,}048} & \textbf{4{,}096} & \textbf{1{,}024} & \textbf{2{,}048} & 4{,}096 & 4{,}096\\
LoRA rank / $\alpha$ / dropout & -- & -- & -- & 128 / 32 / 0.05 & 128 / 32 / 0.05 & 128 / 32 / 0.05\\
DPO $\beta$; DPO:NLL & -- & -- & -- & -- & 0.1; 1.0:0.2 & 0.1; 1.0:0.2\\
Training examples  & 40{,}000 & 40{,}000 & \textbf{47{,}492} & \textbf{98{,}731} & \textbf{28{,}747 pairs} & \textbf{12{,}079 pairs}\\
Gold-answer fallbacks & -- & -- & \textbf{none} & \textbf{1{,}108} & -- & --\\
GPUs               & 8$\times$H100 & 8$\times$H200 & 8$\times$H100 & 8$\times$H200 & 8$\times$H100 & 8$\times$H200\\
GPU-hours (training) & 79.8 & 41.6 & 46.3 & 31.9 & 20.4 & 4.5\\
\bottomrule
\end{tabular}}
\end{table}

\paragraph{S1 data.} 40{,}000 Kimi K3 traces for the USPTO-50k training
reactions, generated on a self-hosted endpoint with reasoning effort
``low'' and an 8{,}192-token limit (14.6M input and 26.5M output tokens,
including hidden reasoning). The teacher sees the starting material,
reagents, conditions (``NA'' for every reaction) and the recorded product,
with no reaction identifiers (prompt in App.~\ref{app:prompts}). The student is
trained on the teacher's returned content (not its internal reasoning field)
inside the thinking block, followed by the recorded product in
\texttt{<answer>} tags.

\paragraph{S2 data.} \emph{8B:} 8 samples per training reaction from
S1 (320{,}000 samples). All finished, exactly correct samples of at most
1{,}024 tokens are kept: 47{,}492 examples covering 19{,}668 reactions
(5{,}691 correct samples are dropped as too long); no gold fallbacks.
\emph{27B:} see App.~\ref{app:s2-27b}.

\paragraph{S3 data.} 16 samples per training reaction from S2
(T\,$=$\,0.7, 8{,}192 new tokens). For each reaction with at least one correct
and one incorrect sample, one correct sample is chosen at random and paired
with the incorrect sample most similar to the target (Tanimoto similarity on
2{,}048-bit Morgan fingerprints, radius 2; invalid SMILES get similarity 0).
Reactions without such a pair are dropped. For 27B, unfinished samples and
sequences over 4{,}096 tokens are removed before pairing (28 removed); for 8B
no such filter is applied (an unfinished sample counts as incorrect, and
DPO truncates to 4{,}096 tokens). One pair is kept per reaction:
28{,}747 of 40{,}000 reactions (71.9\%) yield a pair for 8B
and 12{,}079 (30.2\%) for 27B.

\paragraph{Compute.} Training GPU-hours are in Table~\ref{tab:train-config}.
Sampling for S2 took about 4.2 H100-hours (8B) and 37.6 H200-hours (27B);
sampling for S3 took at most about 8 H100-hours for 8B and 60 H100-hours for 27B.

\subsection{Generation and evaluation protocols}
\label{app:eval}

\begin{table}[h]
\centering\small
\caption{\textbf{Sampling settings.} Seeds: ``fixed'' means the same seed on every request.}
\label{tab:sampling}
\resizebox{\linewidth}{!}{%
\begin{tabular}{lccc}
\toprule
& S2 data & S3 data & Evaluation\\
\midrule
Sampling model       & S1 & S2 & each checkpoint\\
Samples per prompt   & 8 & 16 & 1 (8 for \S\ref{sec:reasoning} probes)\\
Temperature          & 0.9 & 0.7 & 0.6 (base 8B: 0, greedy)\\
Top-$p$ / top-$k$    & 0.95 / 20 & 0.95 / 20 & 0.95 / 20\\
Max new tokens       & 1{,}024 & 8{,}192 & 16{,}384 (8{,}192 for probes)\\
Context length       & 2{,}048 & 16{,}384 (8B) / 10{,}240 (27B) & 16{,}384 (8B) / 20{,}480 (27B)\\
Seed, 27B            & $42+i$ & $42+i$ & $42+i$ for test row $i$\\
Seed, 8B             & none & none & fixed 42 (S1, S3); none (S2); $42+i$ (probes)\\
Thinking enabled     & yes & yes & yes\\
\bottomrule
\end{tabular}}
\end{table}

\paragraph{Frontier models.} Kimi K3, Claude Opus 5 and GPT-6-Astra are
evaluated zero-shot on the same 5{,}000 test reactions with the evaluation
prompt in App.~\ref{app:prompts} (short system prompt), one sample each,
16{,}384 maximum output tokens (for GPT-6-Astra this limit includes hidden
reasoning).
\emph{Kimi K3} (\texttt{moonshotai/Kimi-K3}, self-hosted endpoint, reasoning
enabled at effort ``low''; completed 2026-09-16): 57.4\%.
\emph{Claude Opus 5} (\texttt{claude-opus-5}, Anthropic API, adaptive
thinking, provider-default sampling since the API rejects a temperature;
2026-08-10 to 2026-09-12): 78.7\%.
\emph{GPT-6-Astra} (\texttt{gpt-6-astra}, OpenAI API, default reasoning
effort and sampling; 2026-09-11/12): 63.0\%.
No request is missing from any run; responses with no parsable answer
(9 for Opus 5, 0 for the others) count as incorrect.

\paragraph{Why the probe baselines differ from the headline numbers.}
The \S\ref{sec:reasoning} probes sample each 8B checkpoint eight times per reaction at test time, whereas the headline accuracies in Figure~\ref{fig:accuracy} use a single sample, so their baseline accuracies differ slightly.
\subsection{Prompts}
\label{app:prompts}

\paragraph{Teacher trace generation (S1 data).}
Empty reagent fields are rendered as ``None specified''.
\begin{lstlisting}[style=prompt]
SYSTEM: You are an expert organic chemist with deep knowledge of reaction
mechanisms, reagent roles, and synthetic strategy. Given a reaction described
by SMILES strings, generate a clear, structured reaction trace that would be
useful as training data for a chemistry AI agent focused on retrosynthesis and
forward prediction.

USER:
Starting Material: {starting_material}
Reagents: {reagents}
Reaction Conditions: {conditions}
Product: {product}

Write a reaction trace covering:
1. **Reaction type**: Name and classify this reaction (e.g. Suzuki coupling,
   reductive amination).
2. **Transformation**: Which bonds are broken and which are formed; how the
   carbon skeleton changes.
3. **Mechanism**: Step-by-step electron-pushing mechanism (concise but
   chemically precise).
4. **Reagent roles**: What each reagent/condition does and why it is necessary.
5. **Retrosynthetic note**: The retrosynthetic disconnection that reveals this
   as the key step.

Use standard organic chemistry terminology. Be concise -- aim for 150-300 words.
\end{lstlisting}

\paragraph{S1 training target.} The S1 training prompt is shorter than the
evaluation prompt (it has no instruction block). The same text is used for both models.
\begin{lstlisting}[style=prompt]
SYSTEM: You are an expert organic chemist specializing in the forward
prediction of organic reactions. Your task is to predict the major product of
organic reactions given starting materials and reagents. Return the answer
between <answer> and </answer> tags.

USER: Here are the SMILES of the starting material and reagent(s).
Starting material: {starting_material}
Reagents: {reagents}

ASSISTANT: <think>
{Kimi K3 trace}
</think>

<answer>
{recorded product}
</answer>
\end{lstlisting}

\paragraph{Example teacher trace.} Patent US20120114765A1,
class ``acylation'' (an amide coupling); the teacher used 550 output tokens, including hidden reasoning.
The teacher receives the filled user turn below; the trace it returns is copied
verbatim into the thinking block of the S1 training example, and the recorded
product is the answer. Neither text has been edited.

\begin{lstlisting}[style=trace]
TEACHER INPUT (user turn; system prompt and instructions as above):
Starting Material: O=C(O)c1cc([N+](=O)[O-])c(Sc2c(Cl)cncc2Cl)s1
Reagents: Nc1cccc2cnccc12
Reaction Conditions: NA
Product: O=C(Nc1cccc2cnccc12)c1cc([N+](=O)[O-])c(Sc2c(Cl)cncc2Cl)s1
\end{lstlisting}
\begin{lstlisting}[style=trace]
SYSTEM: You are an expert organic chemist specializing in the forward
prediction of organic reactions. Your task is to predict the major product of
organic reactions given starting materials and reagents. Return the answer
between <answer> and </answer> tags.

USER: Here are the SMILES of the starting material and reagent(s).
Starting material: O=C(O)c1cc([N+](=O)[O-])c(Sc2c(Cl)cncc2Cl)s1
Reagents: Nc1cccc2cnccc12

ASSISTANT: <think>
## Reaction Trace

**1. Reaction type:** Amide bond formation (peptide-type coupling) — a carboxylic acid is condensed with a primary aromatic amine (5-aminoisoquinoline) to give a carboxamide. Classified as a nucleophilic acyl substitution / condensation.

**2. Transformation:**
- **Broken:** The C–OH bond of the thiophene carboxylic acid (the –OH is lost); one N–H bond of the aminoisoquinoline.
- **Formed:** A new C(=O)–N amide bond linking the thiophene carbonyl carbon to the isoquinoline nitrogen. Net loss of H₂O.
- The carbon skeleton is unchanged; the nitro, chloro, and thioether (S-linked dichloropyridine) substituents on the thiophene are spectators.

**3. Mechanism:**
1. The carboxylic acid is activated (in situ by a coupling reagent or via prior conversion to an acid chloride/active ester), making the carbonyl more electrophilic.
2. The lone pair of the aniline-type nitrogen attacks the carbonyl carbon, forming a tetrahedral intermediate (zwitterionic alkoxide).
3. Proton transfer neutralizes the intermediate.
4. Collapse of the tetrahedral intermediate: the C–O bond of the leaving group breaks, the carbonyl reforms, and the activated oxygen group (or OH after activation) departs.
5. Deprotonation of the ammonium nitrogen furnishes the neutral amide.

**4. Reagent roles:**
- **5-Aminoisoquinoline:** nucleophilic coupling partner; its ring nitrogen (aromatic amine) attacks the acyl carbon. Used because the target amide bears this heteroaryl group.
- **Coupling conditions (implied, e.g., EDC/HOBt, SOCl₂, or HATU):** convert the unreactive –OH into a good leaving group; carboxylic acids and amines alone only form a salt and do not condense.

**5. Retrosynthetic note:** Disconnect the C(=O)–N amide bond — the classic amide disconnection. This reveals the thiophene carboxylic acid (acyl synthon) plus 5-aminoisoquinoline (amine synthon), identifying amide coupling as the final key step after the thioether-linked nitrothiophene core is assembled.
</think>

<answer>
O=C(Nc1cccc2cnccc12)c1cc([N+](=O)[O-])c(Sc2c(Cl)cncc2Cl)s1
</answer>
\end{lstlisting}

The trace contains one chemical error that is kept in the training data: under
reagent roles it says the ring nitrogen of 5-aminoisoquinoline attacks the acyl
carbon. The nucleophile is the exocyclic amino group; the product SMILES and the
rest of the trace show this correctly.
\paragraph{Evaluation prompt.} The user turn is identical for all models and
stages. Two system prompts were used: the short one for the 8B students
(all stages and all \S\ref{sec:reasoning} probes) and for the three frontier models; the long
one for the 8B base model, every 27B evaluation (base, S1, S2, S3) and the
8B S2- and S3-data sampling.
\begin{lstlisting}[style=prompt]
USER:
Here are the SMILES of the starting material and reagent(s).
Starting material: {starting_material}
Reagents: {reagent_list}

Predict the single most likely primary product of the chemical reaction in
SMILES format.
The reaction should be chemically reasonable, following typical organic
reactivity principles.
Instructions:
  1. Identify the functional groups in the starting material
  2. Determine which reagent(s) will react with these functional groups
  3. Apply your knowledge of organic reaction mechanisms to predict the major
     product
  4. Consider regioselectivity and stereoselectivity where applicable
  If your proposed mechanism or product seems chemically implausible, revise
  your reasoning and try a different mechanistic path.

Return the answer between <answer> and </answer> tags.
\end{lstlisting}
\begin{lstlisting}[style=prompt]
SYSTEM (short):
You are an expert organic chemist specializing in the forward prediction of
organic reactions.
Your task is to predict the major product(s)
of organic reactions given starting materials and reagents.
\end{lstlisting}
\begin{lstlisting}[style=prompt]
SYSTEM (long):
You are an expert organic chemist specializing in the forward prediction of
organic reactions.
Correctly predicting the product of an organic reaction involves understanding
of several key concepts.
Some of the most important concepts to consider include:

1. Acid-base chemistry: these reactions will proceed from a stronger acid to a
   weaker one.
2. Single displacement reactions: Similarly, these reactions will proceed by a
   more reactive element displacing a less reactive one.
3. Redox reactions: identify the relationship between reagents and their
   potential for reducing/oxidizing specific functional groups on a molecule.
4. Mechanistic analysis: leveraging knowledge of common reaction mechanisms,
   one can infer key functional groups in a molecule and their potential
   relationships to reagents and other key intramolecular functional groups

Your task is to predict the major product(s)
of organic reactions given starting materials and reagents.
\end{lstlisting}

\paragraph{Reasoning-length prompts (\S\ref{sec:reasoning}).}\sloppy
Each is appended to the evaluation prompt after a blank line; the control is
an unmodified resample.
\begin{itemize}[leftmargin=*,itemsep=1pt]
\item \textbf{Commit:} \texttt{Reason directly to the answer. Commit to your first well-supported conclusion and do not revisit earlier steps or explore alternatives.}
\item \textbf{Consider alternatives:} \texttt{Before finalizing, consider at least two plausible product hypotheses and actively check whether any earlier inference could be wrong. Revise if a better interpretation is available.}
\item \textbf{Think more:} \texttt{Work through this carefully and at length: consider the plausible reaction pathways before committing, and check your proposed product against the starting material and reagents.}
\end{itemize}

\paragraph{Feedback prompts (\S\ref{sec:reasoning}).}
Sent as a second user turn after the model's first answer, whatever
that answer was.
\begin{itemize}[leftmargin=*,itemsep=1pt]
\item \textbf{Neutral review (baseline):} \texttt{Review your work above. If you would keep your answer, restate it; if you would change it, give the revised answer instead. Either way, put your final answer between <answer> and </answer> tags.}
\item \textbf{Mild doubt:} \texttt{Please double-check the answer above; there may be an issue with it. Put your final answer between <answer> and </answer> tags.}
\item \textbf{Told wrong:} \texttt{Your answer above is incorrect. Work out where the reasoning went wrong and give the correct product. Put your final answer between <answer> and </answer> tags.}
\item \textbf{Marked wrong (authority):} \texttt{An evaluator marked the answer above as incorrect. Revise it accordingly and give the correct product. Put your final answer between <answer> and </answer> tags.}
\item \textbf{Told correct:} \texttt{Your answer above is correct. Verify it and return the final answer. Put your final answer between <answer> and </answer> tags.}
\end{itemize}
\paragraph{What the model sees in the second turn.} The conversation is
[system, task, first response, feedback]. The Qwen3 chat template removes the
thinking block from earlier assistant turns, so in every feedback condition the
model sees only its previous final answer, not its reasoning trace.
\section{What does each training stage contribute?}
\label{app:stages}

\subsection{The base model does not fail only because of format}
\label{app:fewshot}

\qhead{Is the base model's 1.2\% accuracy just a failure to produce the answer format?}
{\S\ref{sec:stages}: S1 is needed to establish task competence.}
{No. Few-shot examples raise format compliance to 89\% of finished generations but leave accuracy at 1--2\%; over half of the valid answers copy the starting material.}

\begin{table}[h]
\centering\small
\caption{\textbf{Few-shot control on base Qwen3-8B} (USPTO-50k test, $N=5{,}000$, greedy decoding, 16{,}384-token budget, short system prompt). The $k=0$ column is a separate run from the headline base result (1.18\%, long system prompt). The in-context examples are reasoning traces written by Claude Sonnet 4.6 for training reactions.}
\label{tab:fewshot}
\begin{tabular}{lrrr}
\toprule
& $k=0$ & $k=4$ & $k=8$\\
\midrule
Compliance, all generations (\%)   & 60.64 & 65.22 & 67.78\\
Compliance, finished only (\%)     & 75.57 & 88.59 & 89.21\\
Invalid SMILES in tag (\%)         & 19.4  & 8.3 & 8.1\\
Truncated (\%)                     & 19.8  & 26.4 & 24.0\\
Echo rate, \% of valid             & 43.7  & 52.7 & 53.2\\
Top-1 accuracy (\%)                & 1.32  & 1.20 & 2.02\\
\bottomrule
\end{tabular}
\end{table}
The $k=8$ gain over $k=0$ is small but significant (83 vs.\ 48 discordant reactions, McNemar $p=0.003$).

\subsection{Is rejection-sampling SFT (S2) necessary?}
\label{app:skip-s2}

\qhead{Could DPO applied directly to S1 recover the gains of S2 + S3?}
{\S\ref{sec:stages}: rejection sampling contributes beyond preference optimization.}
{Only partly for 8B: skipping S2 reaches 37.1\% vs.\ 42.8\% for the full pipeline (paired difference $-5.7$ points, 95\% CI [$-6.9$, $-4.6$]), recovering about half of the S1$\to$S3 gain. For 27B, skipping S2 does not improve on S1 ($-0.2$ points, $p=0.51$); the full pipeline is 0.8 points above it ($p=0.026$, uncorrected).}

\begin{table}[h]
\centering\small
\caption{\textbf{Skip-S2 ablation.} DPO applied directly to S1, same S3 recipe as the full pipeline for each model (8B: 3{,}000 steps; 27B: 1 epoch). Paired difference = skip-S2 $-$ full pipeline.}
\label{tab:skip-s2}
\begin{tabular}{lcccc}
\toprule
Model & S1 & S1$\to$S2$\to$S3 & S1$\to$S3 (skip S2) & Paired diff.\ [95\% CI], $p$\\
\midrule
Qwen3-8B    & 31.0 & 42.8 & 37.1 & $-5.7$ [$-6.9$, $-4.6$], $p<10^{-20}$\\
Qwen3.8-27B & 79.0 & 79.6 & 78.8 & $-0.8$ [$-1.5$, $-0.1$], $p=0.026$\\
\midrule
DPO pairs, 8B  & & 28{,}747 & 26{,}661 & \\
DPO pairs, 27B & & 12{,}079 & 13{,}698 & \\
\bottomrule
\end{tabular}
\end{table}
The number of available preference pairs does not explain the 8B result:
starting from S1 yields only 7\% fewer pairs than starting from S2
(26{,}661 vs.\ 28{,}747). Skip-S2 is 6.0 points above S1 for 8B
([4.8, 7.3]).

\subsection{Why S2 was rebalanced for the 27B model}
\label{app:s2-27b}

\qhead{The 27B model's S2 differs from the 8B model's. Does that explain why self-training adds nothing at 27B?}
{\S\ref{sec:27b}: a stronger starting point needs less of the pipeline.}
{No. After S1 the 27B model is already correct on 85.8\% of its training samples and solves 65.5\% of training reactions in all 8 samples, so naive rejection sampling would mostly retrain on reactions it already solves. We therefore rebalanced S2 toward reactions it gets only partly right, which gives self-training its best chance. It still adds no measurable accuracy (+0.3 points, $p=0.39$).}

\paragraph{How much is left to learn after S1.}
\begin{table}[h]
\centering\small
\caption{\textbf{Training-set difficulty after S1} (8 samples per training reaction; S2-data sampling settings of Table~\ref{tab:sampling}). A sample counts as correct if it is exactly correct, before any length filter.}
\label{tab:s1-solve}
\begin{tabular}{lcc}
\toprule
& Qwen3-8B S1 & Qwen3.8-27B S1\\
\midrule
Training-set accuracy, per sample (\%)            & 16.6 & 85.8\\
Reactions with $\geq$1 correct sample (\%)        & 51.8 & 97.2\\
Reactions with all 8 correct (\%)                 & 0.3 (116) & 65.5 (26{,}197)\\
Reactions with 1--7 correct (``mixed'', \%)        & 51.5 (20{,}588) & 31.7 (12{,}695)\\
Reactions with 0 correct (\%)                     & 48.2 (19{,}296) & 2.8 (1{,}108)\\
\bottomrule
\end{tabular}
\end{table}
For 8B, the 1{,}024-token filter lowers the share of reactions with a usable
correct sample from 51.8\% to 49.2\% (19{,}668). For 27B every correct sample
fits the 2{,}048-token bound. The 8B training-set accuracy (16.6\%) is below its
test accuracy (31.0\%) because S2 data are sampled at $T=0.9$ with a
1{,}024-token cap.

\paragraph{27B S2 construction.} For reactions with 1--7 usable correct
samples, all are kept (65{,}082 examples). For reactions with 8 of 8
correct, one sample is kept, plus a second for a random subset (32{,}541
``easy replay'' examples; 2:1 mixed to easy). For the 1{,}108 reactions
with no correct sample, the recorded product is given as an answer-only
target. Total: 98{,}731 examples (98.9M tokens) covering all 40{,}000 reactions.
The gold fallbacks are supervision from the dataset labels, not
self-generated; they are 1.1\% of S2 examples.

\paragraph{Recipe selection.} The 27B S2 recipe was chosen after several
variants. One earlier variant (LoRA, 3 epochs, 19{,}096 examples, one shortest
correct trace per reaction with upsampling of weak classes) was evaluated on
the test set (78.2\%) before the final recipe was fixed. The final recipe uses rank-128 LoRA (the same adapter configuration as S3) for one epoch because full fine-tuning is expensive at this scale: a full fine-tuning run on all 274{,}658 correct samples ran at about 0.21 epochs per 5 hours on 8$\times$H200 (about 73 hours for 3 epochs), whereas LoRA processed about 2.3$\times$ more examples per second (7.1 vs.\ 3.1) and one epoch of the final difficulty-mixed dataset (98{,}731 examples) took 4.0 hours. Parameter-efficient fine-tuning is the standard way to adapt models of this size.

\paragraph{The adapter changes the model.} Although S1 and S2 have almost the
same accuracy, 345 test reactions (6.9\%) change outcome between them: 181
become correct and 164 become incorrect.

\paragraph{Base-model knowledge.} Before any training, Qwen3.8-27B reaches
26.3\% top-1. Only 40.8\% of its generations close their thinking block
within 16{,}384 tokens; on those it is 54.2\% accurate (1{,}105 of 2{,}039),
and a further 209 correct answers are stated inside unfinished thinking.

\subsection{Stage-by-stage significance}
\label{app:stage-tests}

\begin{table}[h]
\centering\small
\setlength{\tabcolsep}{4pt}
\caption{\textbf{Paired tests between stages} (McNemar, continuity-corrected; $N=5{,}000$). ``Only A/B'': reactions correct only under A/B. The 8B base model is decoded greedily; all other checkpoints at $T=0.6$.}
\label{tab:stage-tests}
\begin{tabular}{llrrrrrr}
\toprule
Model & Comparison & Acc.\ A & Acc.\ B & Only A & Only B & $p$ & $\Delta$ [95\% CI]\\
\midrule
8B  & base $\to$ S1 & 1.2  & 31.0 & 25 & 1{,}517 & $<10^{-300}$ & $+29.8$ [28.5, 31.1]\\
8B  & S1 $\to$ S2   & 31.0 & 38.9 & 433 & 828 & $1\times10^{-28}$ & $+7.9$ [6.5, 9.3]\\
8B  & S2 $\to$ S3   & 38.9 & 42.8 & 452 & 646 & $6\times10^{-9}$ & $+3.9$ [2.6, 5.2]\\
27B & base $\to$ S1 & 26.3 & 79.0 & 99 & 2{,}735 & $<10^{-300}$ & $+52.7$ [51.2, 54.2]\\
27B & S1 $\to$ S2   & 79.0 & 79.3 & 164 & 181 & 0.39 & $+0.3$ [$-0.4$, 1.1]\\
27B & S2 $\to$ S3   & 79.3 & 79.6 & 69 & 82 & 0.33 & $+0.3$ [$-0.2$, 0.7]\\
27B & S1 $\to$ S3   & 79.0 & 79.6 & 154 & 184 & 0.12 & $+0.6$ [$-0.1$, 1.3]\\
\bottomrule
\end{tabular}
\end{table}
Each configuration was trained once; there are no seed replicates for the
Kimi K3 lineage. Test-set $p$-values therefore do not capture variation between
training runs.

\subsection{Accuracy by stage for Resonance-27B}
\label{app:27b-acc}

\begin{figure}[h]
\centering
\includegraphics[width=\linewidth]{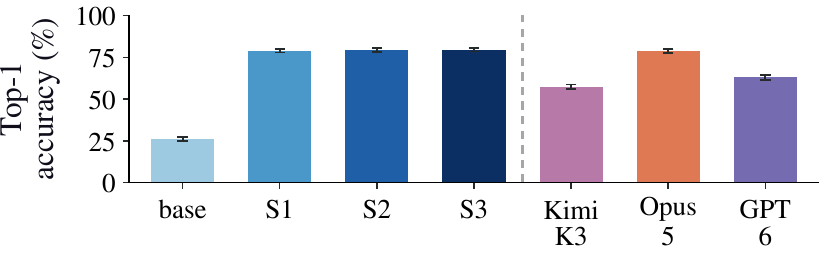}
\caption{\textbf{Top-1 exact-match accuracy on the USPTO-50k test set} ($N=5{,}000$) for Resonance-27B at each stage (base $\rightarrow$ S1 teacher SFT $\rightarrow$ S2 rejection-sampling SFT $\rightarrow$ S3 DPO), and for Kimi K3 (teacher), Claude Opus 5 and GPT-6-Astra, all three evaluated zero-shot. Teacher SFT supplies nearly all of the gain, lifting the model above its teacher, and the two self-training stages change accuracy by less than one point. This is the 27B counterpart of Figure~\ref{fig:accuracy}; the $y$-axis extends to 100\%. Error bars are 95\% Wilson intervals.}
\label{fig:27b-acc}
\end{figure}

\section{Where do the students beat the teacher?}
\label{app:teacher}

\subsection{Accuracy by class for every model and stage}
\label{app:per-class}

\qhead{Which reaction classes drive the aggregate numbers, and where does each model lead?}
{\S\ref{sec:families} and Figure~\ref{fig:reaction_data}.}
{The 8B student leads the teacher on deprotections, reductions and oxidations and trails on the other eight classes; it never learns heterocycle formation ($\le$1\% at every stage). The 27B student (S3) matches or beats Kimi K3 on 9 of 11 classes, all but C--C bond formation (81.5 vs.\ 81.9\%) and FG addition ($n=17$); on heterocycle formation it passes the teacher only after S3 (S1 42.7, S3 48.5, Kimi 44.7\%).}

\begin{table}[h]
\centering\scriptsize
\setlength{\tabcolsep}{3pt}
\caption{\textbf{Top-1 accuracy (\%) by class.} All frontier models zero-shot. $N=5{,}000$.}
\label{tab:per-class}
\begin{tabular}{lrrrrrrrrrr}
\toprule
& & \multicolumn{4}{c}{Qwen3-8B} & \multicolumn{2}{c}{Qwen3.8-27B} & & & \\
\cmidrule(lr){3-6}\cmidrule(lr){7-8}
Class & $n$ & base & S1 & S2 & S3 & S1 & S3 & Kimi K3 & Opus 5 & GPT-6\\
\midrule
Heteroatom alk./aryl. & 1{,}479 & 0.8 & 24.5 & 31.1 & 34.6 & 82.3 & 82.8 & 70.7 & 81.7 & 83.0\\
Acylation             & 1{,}107 & 2.6 & 39.2 & 51.8 & 53.7 & 87.7 & 88.2 & 82.7 & 90.3 & 92.3\\
Deprotections         &   824 & 0.2 & 37.3 & 50.5 & 59.3 & 75.4 & 77.1 & 22.9 & 81.3 & 13.8\\
C--C bond formation   &   536 & 1.1 & 34.9 & 38.4 & 40.5 & 82.6 & 81.5 & 81.9 & 88.2 & 89.0\\
Reductions            &   457 & 0.2 & 31.3 & 33.9 & 43.8 & 67.0 & 65.4 & 2.6 & 41.1 & 2.6\\
Unclassified          &   184 & 0.5 & 14.7 & 14.7 & 15.2 & 66.8 & 67.9 & 20.1 & 54.9 & 16.8\\
FG interconversion    &   127 & 4.7 & 29.1 & 33.1 & 34.6 & 73.2 & 78.0 & 69.3 & 74.0 & 61.4\\
Protections           &   111 & 1.8 & 35.1 & 37.8 & 39.6 & 85.6 & 82.9 & 79.3 & 89.2 & 91.0\\
Heterocycle formation &   103 & 0.0 & 0.0 & 1.0 & 0.0 & 42.7 & 48.5 & 44.7 & 69.9 & 75.7\\
Oxidations            &    55 & 0.0 & 25.5 & 41.8 & 21.8 & 58.2 & 60.0 & 1.8 & 38.2 & 0.0\\
FG addition           &    17 & 0.0 & 0.0 & 5.9 & 0.0 & 29.4 & 52.9 & 58.8 & 41.2 & 64.7\\
\midrule
All                   & 5{,}000 & 1.2 & 31.0 & 38.9 & 42.8 & 79.0 & 79.6 & 57.4 & 78.7 & 63.0\\
\bottomrule
\end{tabular}
\end{table}

\subsection{Are the teacher's low scores real errors?}
\label{app:teacher-errors}

\qhead{A frontier model scoring a few percent on redox suggests its answers may be chemically reasonable but recorded differently. Is the student--teacher gap about chemistry or about dataset conventions?}
{\S\ref{sec:families}: the students lead where the recorded product cannot be read off the prompt.}
{Not about scoring conventions. Ignoring salts, stereochemistry and tautomers raises Kimi K3's reduction accuracy only from 2.6 to 2.8\% and leaves oxidations at 1.8\%. The deficit coincides with prompts that list no reagent: nearly all reduction and oxidation test reactions read ``Reagents: No reagents'', and Kimi K3 then returns an input unchanged in 413 of its 445 reduction errors, i.e.\ it predicts no reaction. The students, trained on the recorded products, almost never do (8B S3: 6 of 257).}

\paragraph{Looser matching.} We rescore every model under progressively
looser equivalence, applied to both prediction and reference: (i) exact
canonical match (the paper's metric); (ii) largest organic fragment only
(removes salts, counter-ions and solvents); (iii) (ii) plus stereochemistry
removed; (iv) (iii) plus tautomer canonicalization (RDKit
\texttt{rdMolStandardize.}\allowbreak\texttt{LargestFragmentChooser},
\texttt{Chem.}\allowbreak\texttt{RemoveStereochemistry},
\texttt{TautomerEnumerator.}\allowbreak\texttt{Canonicalize} with at most
200 tautomers and transforms).

\begin{table}[h]
\centering\small
\caption{\textbf{Accuracy (\%) under looser matching.}}
\label{tab:relaxed}
\resizebox{\linewidth}{!}{%
\begin{tabular}{llcccc}
\toprule
Model & Classes & (i) Exact & (ii) Largest frag. & (iii) + no stereo & (iv) + tautomer\\
\midrule
Kimi K3   & All           & 57.4 & 57.6 & 60.4 & 60.6\\
          & Reductions    & 2.6 & 2.6 & 2.8 & 2.8\\
          & Oxidations    & 1.8 & 1.8 & 1.8 & 1.8\\
          & Deprotections & 22.9 & 22.9 & 23.5 & 23.5\\
          & Protections   & 79.3 & 79.3 & 84.7 & 84.7\\
\midrule
8B S3     & All           & 42.8 & 42.8 & 44.4 & 44.4\\
          & Reductions    & 43.8 & 43.8 & 44.4 & 44.4\\
          & Oxidations    & 21.8 & 21.8 & 21.8 & 21.8\\
          & Deprotections & 59.3 & 59.3 & 60.8 & 60.8\\
          & Protections   & 39.6 & 39.6 & 41.4 & 41.4\\
\midrule
27B S3    & All           & 79.6 & 79.6 & 83.1 & 83.2\\
Opus 5    & All           & 78.7 & 78.8 & 82.7 & 82.9\\
GPT-6     & All           & 63.0 & 63.2 & 65.2 & 65.3\\
\bottomrule
\end{tabular}}
\end{table}
Stereochemistry accounts for 2--4 points for every model; salts and tautomers
are negligible. No class-level teacher deficit closes.

\paragraph{Where the teacher's errors come from.} 450 of 457 reduction test
prompts, all 55 oxidation prompts and 702 of 824 deprotection prompts read
``Reagents: No reagents'': the reducing, oxidizing or deprotecting agent is not
in the dataset. Table~\ref{tab:error-cats} splits each model's incorrect
predictions automatically. A prediction ``repeats an input'' if, after
canonicalization, it consists only of molecules from the prompt (the starting
material or a listed reagent), i.e.\ the model predicts that no reaction occurs.

\begin{table}[h]
\centering\small
\caption{\textbf{Incorrect predictions that repeat an input} (count / all incorrect predictions in the class).}
\label{tab:error-cats}
\begin{tabular}{lccc}
\toprule
Model & Reductions & Oxidations & Deprotections\\
\midrule
Kimi K3  & 413 / 445 & 50 / 54 & 577 / 635\\
GPT-6    & 440 / 445 & 55 / 55 & 687 / 710\\
Opus 5   & 10 / 269 & 1 / 34 & 0 / 154\\
8B S3    & 6 / 257 & 0 / 43 & 7 / 335\\
\bottomrule
\end{tabular}
\end{table}
Kimi K3 and GPT-6 answer ``no reaction'' when no reagent is listed; Opus 5 and
the students instead propose a transformation. The students' remaining errors are
mostly a different wrong product (8B S3: 226 of 257 reduction errors), with some
invalid SMILES (23 of 257).

\subsection{Do students and teacher solve the same reactions?}
\label{app:overlap}

\qhead{Similar aggregate accuracy can hide very different behavior. How much do the correct sets overlap?}
{\S\ref{sec:families}: students and teacher succeed on different chemistry.}
{The 8B student and Kimi K3 agree only slightly beyond chance ($\kappa=0.11$): each solves many reactions the other misses (765 vs.\ 1{,}497), and a union oracle reaches 72.7\%. The 27B student covers most of the teacher's successes; only 288 reactions (5.8\%) are solved by the teacher alone.}

\begin{table}[h]
\centering\small
\caption{\textbf{Paired outcomes against the teacher.} $N=5{,}000$; counts with \% of reactions.}
\label{tab:overlap}
\begin{tabular}{lcc}
\toprule
& 8B S3 vs.\ Kimi K3 & 27B S3 vs.\ Kimi K3\\
\midrule
Both correct              & 1{,}375 (27.5) & 2{,}584 (51.7)\\
Only student correct      & 765 (15.3) & 1{,}396 (27.9)\\
Only teacher correct      & 1{,}497 (29.9) & 288 (5.8)\\
Both wrong                & 1{,}363 (27.3) & 732 (14.6)\\
McNemar $\chi^2$ ($p$)    & 236.2 ($3\times10^{-53}$) & 727.7 ($3\times10^{-160}$)\\
Jaccard overlap of correct sets & 0.378 & 0.605\\
Cohen's $\kappa$          & 0.114 & 0.261\\
Union-oracle accuracy (\%) & 72.7 & 85.4\\
\bottomrule
\end{tabular}
\end{table}
\section{How does the student reason?}
\label{app:reasoning}

\subsection{Measuring backtracking}
\label{app:lexicon}

\begin{figure}[h]
\centering
\includegraphics[width=0.9\linewidth]{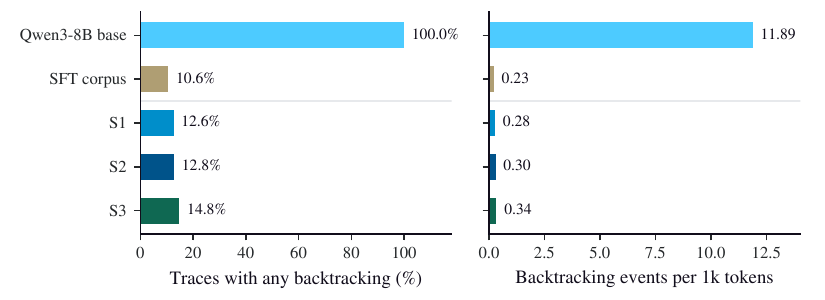}
\caption{\textbf{The student inherits the corpus's reasoning style, not the base model's.} Share of traces with any backtracking (left) and events per 1k output tokens (right), default prompt.}
\label{fig:style}
\end{figure}

\begin{figure}[h]
\centering
\includegraphics[width=0.95\linewidth]{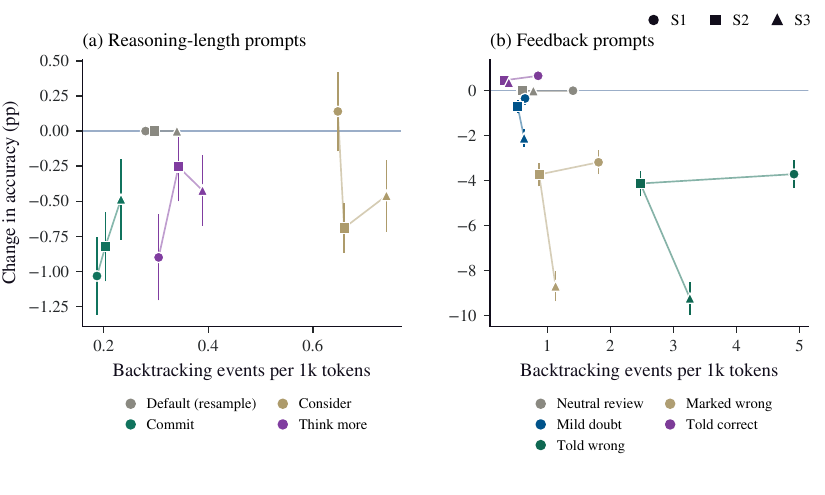}
\caption{\textbf{Prompts raise backtracking without raising accuracy.} Change in accuracy against backtracking events per 1k output tokens for (a) reasoning-length prompts and (b) feedback prompts, by stage. ``Consider alternatives'' raises backtracking about 2.4$\times$ (0.27$\to$0.66 per 1k at S1) with no accuracy gain; ``told wrong'' raises it 8--18$\times$ over the default resample (3.5--4.1$\times$ over the neutral review) and produces the largest losses.}
\label{fig:a2}
\end{figure}

\qhead{Is counting backtracking phrases a valid measure of reconsidering?}
{\S\ref{sec:reasoning} and Figs.~\ref{fig:style}--\ref{fig:a2}.}
{It is a surface measure. The lexicon was fixed before any intervention was run, but it has not yet been validated against hand labels, so we use it only to compare conditions, not as a count of real revisions.}

\paragraph{Lexicon.} We count case-insensitive, word-bounded matches of a
fixed list of phrases inside the thinking block only. The list was frozen
(definition 2026-09-12.1) before any intervention was generated:
\begin{itemize}[leftmargin=*,itemsep=1pt]
\item \emph{Reversal:} wait; hold on; hmm; oops; scratch that; that/this/the above is (or was, would be) wrong/incorrect/not right/a mistake; I made a mistake/error; no, that is wrong; I was wrong/mistaken; on second thought; actually.
\item \emph{Revision:} instead (of); rather than; let me re-\ldots/try again/start over/redo/rethink/reconsider/revisit; reconsider(ing); revise/revising/revised/revision; correction; on closer/further inspection/reflection/thought; going back; back up; reverse/reversing.
\item \emph{Branching:} alternatively; another possibility/option/route/candidate/way; other possibilities/options; a second possibility/option; option b--z; could also be; might also be; or it could/might be; two possibilities; both are/seem plausible; on the other hand.
\item \emph{Markers:} \texttt{<candidate>}, \texttt{<reconsideration>}, \texttt{<revision>}, \texttt{<alternative>} tags.
\end{itemize}
Rates are reported as events per 1{,}000 output tokens (the whole generation,
counted with the Qwen3-8B tokenizer; for the Kimi K3 corpus, the trace text) and
as the share of traces with at least one event. ``Wait'' accounts for 87\% of
reversal matches; we also computed every rate without it.

\subsection{Where the student's reasoning style comes from}
\label{app:style}

\qhead{Does the student reason like its base model or like its training data?}
{\S\ref{sec:reasoning}: the student inherits the teacher corpus's single-pass style.}
{Like its training data. The base model backtracks in every trace (11.9 events per 1k tokens); the Kimi K3 corpus rarely does (10.6\% of traces, 0.23 per 1k); after SFT the student matches the corpus (0.27 per 1k), and self-training barely changes this.}

\begin{table}[h]
\centering\small
\caption{\textbf{Backtracking by source} (default prompt). The base model is sampled with the evaluation settings ($T=0.6$); its generations average 3{,}934 tokens.}
\label{tab:style}
\resizebox{\linewidth}{!}{%
\begin{tabular}{lcc}
\toprule
& Traces with any backtracking (\%) & Events per 1k tokens\\
\midrule
Qwen3-8B base (before training)    & 100  & 11.9\\
Kimi K3 SFT corpus (training data, $n=40{,}000$) & 10.6 & 0.23\\
S1                                 & 12.6 & 0.27\\
S2                                 & 13.6 & 0.32\\
S3                                 & 14.6 & 0.33\\
\bottomrule
\end{tabular}}
\end{table}
Without ``wait'', the base model still produces 6.2 events per 1k tokens.

\paragraph{Caveat.} The corpus's lack of backtracking has at least two
possible causes: the teacher saw the recorded product, and the
generation prompt asked for a concise structured trace with fixed
headings (App.~\ref{app:prompts}). We do not separate these.

\subsection{Prompts that change reasoning length}
\label{app:length-prompts}

\qhead{Does asking the model to commit, consider alternatives or think more change accuracy?}
{\S\ref{sec:reasoning} (naive test-time scaling).}
{No prompt improves accuracy at any stage. The effects are small (at most 1.03 points) but mostly negative: moving the model away from its trained default in either direction costs a little.}

\paragraph{Design.} Each prompted condition is paired with an unprompted
resample of the same model on the same 5{,}000 reactions, 8 draws per
reaction, with independent per-reaction seeds (per-draw accuracies are
consistent with independent draws, $\chi^2$ $p=0.26$--$1.00$). The per-reaction
mean difference is tested across reactions (App.~\ref{app:stats}).

\begin{table}[h]
\centering\small
\setlength{\tabcolsep}{4pt}
\caption{\textbf{Change in top-1 accuracy (points) vs.\ unprompted resample}, with $p$-values. Baseline accuracy S1/S2/S3: 29.0 / 38.8 / 41.3\%. Mean reasoning tokens in brackets.}
\label{tab:length-prompts}
\resizebox{\linewidth}{!}{%
\begin{tabular}{lccc}
\toprule
Prompt & S1 & S2 & S3\\
\midrule
Default (resample) & 0 \;[488] & 0 \;[469] & 0 \;[475]\\
Commit to first answer & $-1.03$ ($p<0.001$) [451] & $-0.82$ ($p<0.001$) [432] & $-0.49$ ($p=0.026$) [437]\\
Consider alternatives  & $+0.14$ ($p=0.56$) [516]     & $-0.69$ ($p=0.003$) [496] & $-0.46$ ($p=0.042$) [505]\\
Think more             & $-0.90$ ($p<0.001$) [651] & $-0.26$ ($p=0.28$) [608]     & $-0.42$ ($p=0.062$) [612]\\
\bottomrule
\end{tabular}}
\end{table}

\paragraph{Multiple comparisons.} Across the 9 cells, 6 are significantly
negative at $p<0.05$ and none is significantly positive. Under a
Holm--Bonferroni correction at $\alpha=0.05$, 4 remain significant:
S1 commit ($p=2.4\times10^{-5}$), S1 think more ($2.4\times10^{-4}$),
S2 commit ($2.5\times10^{-4}$) and S2 consider ($2.7\times10^{-3}$); the next
smallest, S3 commit ($p=0.026$), exceeds its threshold of $0.05/5$.

\paragraph{Format failures versus chemistry.} At S1, the losses under
``commit'' and ``think more'' come from answers that fail to parse. Restricted
to parsable answers, S1 commit is $+0.98$ ($p=0.07$) and S1 think more is
$+1.15$ ($p=0.038$), while S1 consider is $-1.02$ ($p=0.052$). At S2 and S3 the
differences stay negative or non-significant.

\subsection{Response to feedback}
\label{app:feedback}

\qhead{When told its answer is wrong or right, regardless of correctness, does the model weigh the feedback or simply comply?}
{\S\ref{sec:reasoning}: negative feedback makes the model abandon correct answers, most sharply after preference optimization.}
{Negative feedback costs 3.2--3.7 points at S1 and 8.7--9.2 at S3; ``told correct'' is the only intervention that helps (by under one point). Told it is wrong, the model abandons 22--32\% of its correct answers but rescues only about 6\% of its wrong ones: it treats negative feedback as an instruction to change the answer, not as evidence.}

\paragraph{Design.} The model's first answer is returned with one of the
feedback messages in App.~\ref{app:prompts}, whatever that answer's
correctness; the model sees its previous answer but not its reasoning. Each
condition is paired with a neutral ``review your answer'' turn on the same
reactions and first answers, 8 draws per reaction.

\begin{table}[h]
\centering\small
\caption{\textbf{Change in top-1 accuracy (points) vs.\ neutral review}, 8 draws per reaction. Baseline accuracy after neutral review S1/S2/S3: 28.0 / 38.3 / 41.1\%. All cells remain significant under Holm--Bonferroni at $\alpha=0.05$.}
\label{tab:feedback}
\begin{tabular}{lccc}
\toprule
Feedback & S1 & S2 & S3\\
\midrule
Mild doubt               & $-0.34$ ($p=0.020$) & $-0.71$ ($p<10^{-5}$) & $-2.10$ ($p<10^{-26}$)\\
Told wrong               & $-3.71$ ($p<10^{-31}$) & $-4.13$ ($p<10^{-45}$) & $-9.22$ ($p<10^{-134}$)\\
Marked wrong (authority) & $-3.18$ ($p<10^{-31}$) & $-3.73$ ($p<10^{-43}$) & $-8.69$ ($p<10^{-140}$)\\
Told correct             & $+0.66$ ($p<10^{-11}$) & $+0.47$ ($p<10^{-7}$) & $+0.36$ ($p<10^{-3}$)\\
\midrule
\multicolumn{4}{l}{\emph{``Told wrong'' loss as a share of baseline accuracy}}\\
                         & 13\% & 11\% & 22\%\\
\bottomrule
\end{tabular}
\end{table}

\paragraph{Answer flips.} Changes in accuracy hide two opposite flows.
Table~\ref{tab:flips} separates them, measured against the model's
\emph{first} answer (before any second turn), so that the effect of the
neutral review itself is visible. First-answer accuracy is 28.7 / 38.9 / 41.6\%
at S1/S2/S3.

\begin{table}[h]
\centering\small
\setlength{\tabcolsep}{4pt}
\caption{\textbf{Answer flips relative to the first answer} (\%, mean over 8 draws). C$\to$I: share of initially correct answers that end incorrect. I$\to$C: share of initially incorrect answers that end correct. A fresh resample is the no-feedback reference.}
\label{tab:flips}
\begin{tabular}{lcccccc}
\toprule
& \multicolumn{2}{c}{S1} & \multicolumn{2}{c}{S2} & \multicolumn{2}{c}{S3}\\
\cmidrule(lr){2-3}\cmidrule(lr){4-5}\cmidrule(lr){6-7}
Condition & C$\to$I & I$\to$C & C$\to$I & I$\to$C & C$\to$I & I$\to$C\\
\midrule
Fresh resample  & 41.8 & 17.2 & 28.0 & 17.7 & 25.1 & 17.4\\
Neutral review  & 5.7 & 1.3 & 3.2 & 1.1 & 3.6 & 1.8\\
Mild doubt      & 10.6 & 2.8 & 7.8 & 2.9 & 11.7 & 3.9\\
Told wrong      & 31.0 & 6.3 & 21.5 & 6.0 & 32.3 & 6.4\\
Marked wrong    & 26.6 & 5.2 & 19.9 & 5.6 & 30.1 & 5.8\\
Told correct    & 2.5 & 0.9 & 1.9 & 1.0 & 2.6 & 1.7\\
\bottomrule
\end{tabular}
\end{table}
A second turn is far more stable than an independent resample. Relative to the
neutral review, ``told wrong'' raises C$\to$I about 5-fold at S1 and about
9-fold at S3, which is the direct measure of the growing fragility.

\paragraph{What backtracking does not explain.} Across stages, the size of
the ``told wrong'' loss does not track the amount of backtracking: S3
loses the most accuracy while backtracking less than S1
(3.3 vs.\ 4.9 events per 1k). Backtracking counts are therefore
consistent with ``more reconsidering does not help'', but they do not
explain why S3 is the most fragile; the flip rates
(Table~\ref{tab:flips}) are the more direct measure.

\subsection{Do structure changes within a trace reach the right product?}
\label{app:revisions}

\qhead{When the structures the model writes down change during a trace, do the changes move toward the recorded product?}
{\S\ref{sec:reasoning} and \S\ref{sec:conclusions}: the model produces the look of reconsidering without its useful part.}
{Few traces (7--11\%) contain two or more distinct structures, so there is little change to measure. Among changes, reaching the recorded product (wrong$\to$right) is more common than leaving it (right$\to$wrong) at every stage and prompt, but most changes are between two structures that are not the recorded product (72--82\%).}

\paragraph{Method.} We extract every SMILES-like string inside the thinking
block of the first draw that RDKit parses and that is not a component of the
starting material or reagents, and canonicalize it with RDKit. Structures are
compared by canonical SMILES, so the same molecule written two ways is not
counted as a change. A \emph{change} is a step from one extracted structure to
a different one, and each change is classified by comparing both structures
with the recorded product. Extracted structures include intermediates and
by-products as well as candidate products, so ``wrong'' here means ``not the
recorded product'': wrong$\to$wrong changes include steps through a mechanism,
and wrong$\to$right means the trace reaches the recorded product. Consistent
with this, among traces with at least one change the final answer equals the
last extracted structure in only 36--69\% of cases.

\begin{table}[h]
\centering\small
\caption{\textbf{Structure changes within the trace} (RDKit-canonicalized SMILES; first draw). Columns 3--4: \% of traces; columns 5--7: \% of change events; column 8: \% of traces with $\ge$1 change.}
\label{tab:revisions}
\resizebox{\linewidth}{!}{%
\begin{tabular}{llcccccc}
\toprule
Stage & Condition & \makecell{Any\\candidate} & \makecell{$\geq$1\\change} & \makecell{Wrong\\$\to$right} & \makecell{Right\\$\to$wrong} & \makecell{Wrong\\$\to$wrong} & \makecell{Final answer\\= last cand.}\\
\midrule
S1 & Default               & 64.0 & 8.0 & 16.8 & 4.0 & 79.2 & 68.4\\
   & Consider alternatives & 58.2 & 7.8 & 17.0 & 1.4 & 81.6 & 68.9\\
   & Think more            & 64.3 & 9.6 & 18.2 & 3.3 & 78.6 & 60.7\\
   & Told wrong            & 51.4 & 11.1 & 13.3 & 8.5 & 78.2 & 38.7\\
\midrule
S3 & Default               & 58.5 & 7.9 & 25.9 & 2.2 & 71.9 & 62.9\\
   & Consider alternatives & 54.6 & 7.0 & 26.3 & 2.1 & 71.7 & 64.8\\
   & Think more            & 56.7 & 8.4 & 22.5 & 3.0 & 74.5 & 59.0\\
   & Told wrong            & 49.8 & 7.8 & 16.4 & 9.4 & 74.2 & 36.0\\
\bottomrule
\end{tabular}}
\end{table}

\subsection{Sampling several answers per reaction}
\label{app:sampling}

\qhead{Does self-training teach new reactions, or sharpen answers the model could already reach?}
{\S\ref{sec:reasoning}: self-training mostly sharpens.}
{Mostly sharpens. From S1 to S3, single-sample accuracy rises 12.9 points but pass@9 rises only 4.1 (63.8$\to$67.9\%). Plurality voting over nine S1 samples matches or exceeds one S3 sample (46.1 vs.\ 41.6\% overall, and in 8 of the 10 classes with $n\ge40$); heterocycle formation stays near zero at every $k$ for the 8B model (pass@9 $\le$1.9\%).}

\begin{table}[h]
\centering\small
\caption{\textbf{Plurality@$k$ and pass@$k$ (\%), Qwen3-8B, all 5{,}000 reactions.} Nine draws per reaction (the first answer plus 8 fresh resamples); pass@$k$ is the empirical share of reactions with a correct answer among the first $k$ draws.}
\label{tab:plurality}
\begin{tabular}{llccccc}
\toprule
Stage & Metric & $k=1$ & $k=3$ & $k=5$ & $k=7$ & $k=9$\\
\midrule
S1 & plurality@$k$ & 28.7 & 35.4 & 41.3 & 44.2 & 46.1\\
   & pass@$k$      & 28.7 & 47.5 & 55.8 & 60.3 & 63.8\\
S2 & plurality@$k$ & 38.9 & 44.6 & 48.5 & 50.6 & 51.7\\
   & pass@$k$      & 38.9 & 55.0 & 61.2 & 65.0 & 67.4\\
S3 & plurality@$k$ & 41.6 & 46.9 & 50.0 & 52.1 & 54.0\\
   & pass@$k$      & 41.6 & 56.9 & 62.5 & 65.7 & 67.9\\
\bottomrule
\end{tabular}
\end{table}
The exceptions to ``nine S1 samples beat one S3 sample'' are oxidations
(49.1 vs.\ 50.9\%) and heterocycle formation (0 vs.\ 0\%).

\section{Statistical methods}
\label{app:stats}

\paragraph{Comparisons between models or stages.} All models are evaluated
on the same 5{,}000 reactions, so comparisons use paired tests. For one
sample per reaction (App.~\ref{app:stages} and \ref{app:teacher}) we use
McNemar's test with continuity correction on the discordant pairs and report
paired Wald 95\% confidence intervals on the accuracy difference.

\paragraph{Repeated draws.} For the \S\ref{sec:reasoning} probes with 8 draws per reaction,
draws of the same reaction are not independent. We average within each
reaction and test the per-reaction mean difference across reactions with a
two-sided normal ($z$) test, with standard error SD$/\sqrt{n}$ over the
$n=5{,}000$ reactions. Standard errors are therefore clustered by reaction
(0.22--0.25 points for the reasoning-length prompts).

\paragraph{Single-model intervals.} Error bars on single-model accuracy are
95\% Wilson score intervals. They describe uncertainty in one model's
accuracy and are not used to compare models; overlapping Wilson intervals
do not imply equivalence.

\paragraph{Multiple comparisons.} Within each family of related tests
(the 9 reasoning-length cells; the 12 feedback cells) we report
Holm--Bonferroni-adjusted significance alongside raw $p$-values.

\paragraph{Training variance.} Each configuration was trained once. Test-set
$p$-values describe uncertainty over reactions and sampling, not variation
between training runs.

\end{document}